\documentclass[10pt,twocolumn,letterpaper]{article}

\usepackage{iccv}
\usepackage{times}
\usepackage{epsfig}
\usepackage{graphicx}
\usepackage{amsmath}
\usepackage{amssymb}
\usepackage{enumitem}
\usepackage{siunitx}
\usepackage{color}
\usepackage{multirow}
\usepackage{booktabs}
\usepackage{caption}
\usepackage{subcaption}
\usepackage{tabularx}
\usepackage{float}
\usepackage{authblk}

\usepackage[pagebackref=true,breaklinks=true,letterpaper=true,colorlinks,bookmarks=false]{hyperref}

\iccvfinalcopy 

\newcommand{\Yao}[1]{}
\newcommand{\ZK}[1]{}
\newcommand{\Steve}[1]{}

\def\iccvPaperID{787} 

\begin{document}

\title{TS-LSTM and Temporal-Inception: \\ Exploiting Spatiotemporal Dynamics for Activity Recognition}





\newcommand*\samethanks[1][\value{footnote}]{\footnotemark[#1]}

\author[1]{Chih-Yao Ma\thanks{equal contribution}}
\author[1]{Min-Hung Chen\samethanks}
\author[2]{Zsolt Kira}
\author[1]{Ghassan AlRegib}
\affil[1]{Georgia Institute of Technology}
\affil[2]{Georgia Tech Research Institution}
\affil[ ]{\small\textit {\{cyma, cmhungsteve, zkira, alregib\}@gatech.edu}}

\setcounter{Maxaffil}{0}
\renewcommand\Affilfont{\itshape\small}

\maketitle

\begin{abstract}
    Recent two-stream deep Convolutional Neural Networks (ConvNets) have made significant progress in recognizing human actions in videos. Despite their success, methods extending the basic two-stream ConvNet have not systematically explored possible network architectures to further exploit spatiotemporal dynamics within video sequences. Further, such networks often use different baseline two-stream networks. Therefore, the differences and the distinguishing factors between various methods using Recurrent Neural Networks (RNN) or convolutional networks on temporally-constructed feature vectors (Temporal-ConvNet) are unclear. In this work, we first demonstrate a strong baseline two-stream ConvNet using ResNet-101. We use this baseline to thoroughly examine the use of both RNNs and Temporal-ConvNets for extracting spatiotemporal information. Building upon our experimental results, we then propose and investigate two different networks to further integrate spatiotemporal information: 1) temporal segment RNN and 2) Inception-style Temporal-ConvNet. We demonstrate that using both RNNs (using LSTMs) and Temporal-ConvNets on spatiotemporal feature matrices are able to exploit spatiotemporal dynamics to improve the overall performance. However, each of these methods require proper care to achieve state-of-the-art performance; for example, LSTMs require pre-segmented data or else they cannot fully exploit temporal information. Our analysis identifies specific limitations for each method that could form the basis of future work. Our experimental results on UCF101 and HMDB51 datasets achieve state-of-the-art performances, 94.1\% and 69.0\%, respectively, without requiring extensive temporal augmentation.  
\end{abstract}

\begin{figure*}[!htbp]
\begin{center}
   \includegraphics[width=0.85\linewidth]{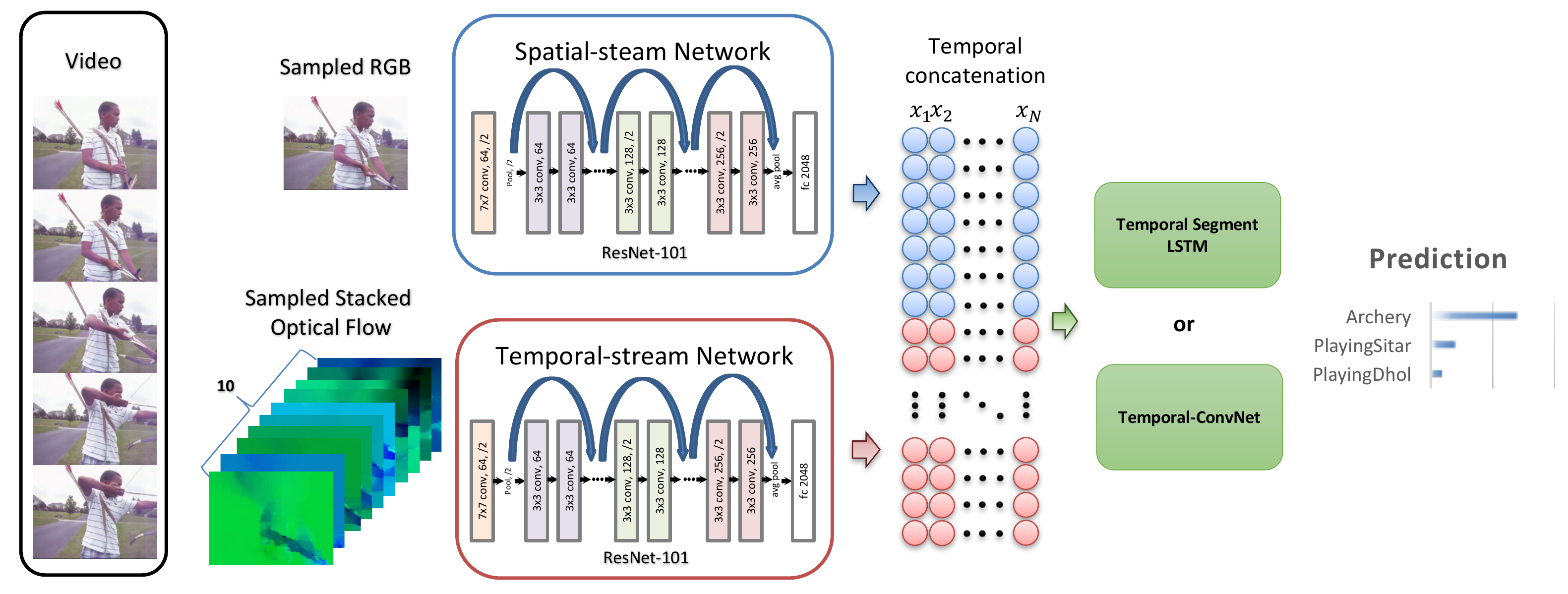}
\end{center}
   \caption{\textbf{Overview of the proposed framework.} Spatial and temporal features were extracted from a two-stream ConvNet using ResNet-101 pre-trained on ImageNet, and fine-tuned for single-frame activity prediction.
   Spatial and temporal features are concatenated and temporally-constructed into feature matrices. The constructed feature matrices are then used as input to both of our proposed methods: Temporal Segment LSTM (TS-LSTM) and Temporal-Inception.
   }
\label{fig:overview}
\end{figure*}

\section{Introduction}
Human action recognition is a challenging task and has been researched for years. Compared to single-image recognition, the temporal correlations between image frames of a video provide additional motion information for recognition. At the same time, the task is much more computationally demanding since each video contains hundreds of image frames that need to be processed individually. 

Encouraged by the success of using Convolutional Neural Networks (ConvNets) on still images, many researchers have developed similar methods for video understanding and human action recognition \cite{simonyan2014two,yue2015beyond,feichtenhofer2016convolutional,zha2015exploiting,sun2015human,tran2015learning}. 
Most of the recent works were inspired by two-stream ConvNets proposed by Simonyan et al. \cite{simonyan2014two}, which incorporate spatial and temporal information extracted from RGB and optical flow images. These two image types are fed into two separate networks, and finally the prediction score from each of the streams are fused. \ZK{Changed: Paragraph separated. Otherwise it was one long block across entire half-page} \Yao{I see. Thank you!}

However, traditional two-stream ConvNets are unable to exploit the most critical component in action recognition, e.g. visual appearance across both spatial and temporal streams and their correlations are not considered. Several works have explored the stronger use of spatiotemporal information, typically by taking frame-level features and integrating them using Long short-term memory (LSTM) cells, temporal feature pooling \cite{yue2015beyond,feichtenhofer2016convolutional} and temporal segments \cite{WangXWQLTV16}. However, these works typically try individual methods with little analysis of whether and how they can successfully use temporal information. Furthermore, each individual work uses different networks for the baseline two-stream approach, with varied performance depending on training and testing procedure as well as the optical flow method used. Therefore it is unclear how much improvement resulted from a different use of temporal information. 

In this paper, we would like to answer the question: given the spatial and motion features representations over time \ZK{Although optical flow is present too} \Yao{I see your concern. Changed to 'spatial and motion'. Hope it's more clear'}, what is the best way to exploit the temporal information? We thoroughly evaluate design choices with respect to a baseline two-stream ConvNet and two proposed methods for exploiting spatiotemporal information. We show that both can achieve high accuracy individually, and our approach using LSTMs with temporal segments improves upon the current state of the art. 

In this work, we aim to investigate several different ways to model the temporal dynamics of activities with feature representations from image appearance and motion. We encode the spatial and temporal features via a high-dimensional feature space, examining various training practices for developing two-stream ConvNets for action recognition. Using this and other related work as a baseline, we then make the following contributions:
\begin{enumerate}
   \item Temporal Segment LSTM (TS-LSTM): we revisit the use of LSTMs to fuse high-level spatial and temporal features to learn hidden features across time. We adapt the temporal segment method by Wang et al.~\cite{WangXWQLTV16} and exploit it with LSTM cells. We show that directly using LSTM performed only similar to naive temporal pooling methods, e.g. mean or max pooling, and the integration of both yields better performance. 
   \item
   Temporal-ConvNet: We propose to use stacked temporal convolution kernels to explore temporal information at multiple scales. The proposed architecture can be extended to an Inception-style Temporal-ConvNet.
   We show that by properly exploiting the temporal information, Temporal-Inception can achieve state-of-the-art performance even when taking feature vectors as inputs (i.e. without using feature maps).
\end{enumerate}

By thoroughly exploring the space of architectural designs within each method, we clarify the contribution of each decision and highlight their implication in terms of current limitations of methods such as LSTMs to fully exploit temporal information without manipulating its inputs\ZK{Depending on space, we may want to just summarize a set of conclusions one can make based on the experiments, for both methods. Space is an issue though.}. Our approaches are implemented using Torch7 \cite{torch} and are publicly available at \href{https://github.com/chihyaoma/Activity-Recognition-with-CNN-and-RNN}{GitHub}. 

\section{Related work}

Many of the action recognition methods extract high-dimensional features that can be used within a classifier. These features can be hand-crafted~\cite{wang2011action,wang2013action,peng2016bag,wang2009evaluation} or learned, and in many instances, frame-level features are then combined in some form. 


\textbf{3D ConvNets.} 
The early work from Karpathy et al.~\cite{karpathy2014large} stacked consecutive video frames and extended the first convolutional layer to learn the spatiotemporal features while exploring different fusion approaches, including early fusion and slow fusion. Another proposed method, C3D~\cite{tran2015learning}, took this idea one step further by replacing all of the 2D convolutional kernels with 3D kernels at the expense of GPU memory. 
To avoid high complexity when training 3D convolutional kernels, Sun et al.~\cite{sun2015human} factorize the original 3D kernels into 2D spatial and 1D temporal kernels and achieve comparable performance.
\Yao{I removed a paragraph here, because I think \cite{sun2015human} also used different kernels. Steve, can you double-check?}
Instead of using only one layer like \cite{sun2015human},
we demonstrate that multiple layers can extract temporal correlations at different time scales and provide better capability to distinguish different types of actions.
\Steve{Yes, they also have different kernel sizes. I revised the paragraph.}

\textbf{ConvNets with RNNs.} 
Instead of integrating temporal information via 3D convolutional kernels, Donahue et al.~\cite{donahue2015long} fed spatial features extracted from each time step to a recurrent network with LSTM cells. In contrast to the traditional models which can only take a fixed number of temporal inputs and have limited spatiotemporal receptive fields, the proposed Long-term Recurrent Convolutional Networks (LRCN) can directly take variable length inputs and learn long-term dependencies. 

\textbf{Two-stream ConvNets.} 
Another branch of research in action recognition extracts temporal information from traditional optical flow images\ZK{Is this really another branch, since above methods typically combine/use two-stream no?} \Yao{\cite{sun2015human} is kind of in between, but I decided to mention it along with C3D because I think it's more closely related. So, yes, it's another branch.}. This approach was pioneered by \cite{simonyan2014two}. The proposed two-stream ConvNets demonstrated that the stacked optical flow images solely can achieve comparable performance despite the limited training data. 
Currently, the two-stream ConvNet is the most popular and effective approach for action recognition. A few works have proposed to extend the approach. Ng et al. \cite{yue2015beyond} take advantage of both two-stream ConvNets and LRCN, in which not only the spatial features are fed into the LSTM but also the temporal features from optical flow images. 
Our work shows that combining a ConvNet with vanilla LSTM results in limited performance improvement when there are not many temporal variances. 
Feichtenhofer et al.~\cite{feichtenhofer2016convolutional} proposed to fuse the spatiotemporal streams via 3D convolutional kernels and 3D pooling. Our proposed Temporal-ConvNet is different because we only use the feature vector representations instead of feature maps. \ZK{Should this be back in the 3D section? Seems redundant with explanation of why our convolutional method is different back in the 3D ConvNets section. Could save some space.} \Yao{\cite{feichtenhofer2016convolutional} use two-stream as base. \cite{sun2015human} doesn't. But I do agree that it is redundant. Also, because I think the reason we claimed to be different from \cite{sun2015human} is incorrect, I removed it.}
We also show that by properly leveraging temporal information, our proposed Temporal-Inception can achieve state-of-the-art results only using feature vector representations. 

Similar to the above works, many other approaches exist. Wang et al.~\cite{WangXWQLTV16} proposed the temporal segment network (TSN), which divides the input video into several segments and extracts two-stream features from randomly selected snippets.
\cite{WangSWVH16}
incorporate semantic information into the two-stream ConvNets.
Zhu et al.~\cite{Zhu_2016_CVPR} propose a key volume mining deep framework to identify key volumes that are associated with discriminative actions. 
Wang et al.~\cite{Wang_2016_CVPR} proposed to use a two-stream Siamese network to model the transformation of the state of the environment.


In this work, we aim to create a common and strong baseline two-stream ConvNet and extend the two-stream ConvNet to provide in-depth analysis of design decisions for both RNN and Temporal-ConvNet. We demonstrate that both methods can achieve state of the art but proper care
must be given. For instance, LSTMs require pre-segmented data or else they cannot fully exploit temporal information. Our analysis identifies specific limitations for each method that could form the basis of future work. \ZK{This paragraph doesn't explain what we do differently; it explains what is pretty much in common with above related work (exploiting spatio-temporal information better than two-stream, using CNNs/LSTMs for this, etc. The description is more of what LSTMs and CNNs do. Need to replace with key differentiators: In-depth analysis of design decisions using a common, strong, two-stream baseline, demonstration that both methods can achieve state of art but with some understanding/manipuation of architectures, highlight of what analysis means in terms of limitations of methods such as LSTMs} \Yao{Totally agree. Rewritten.}


In the following sections, we first discuss our proposed approach in section \ref{sec:approach}. In each of the subsections, the intuition and reasoning behind our approaches will be elaborated upon. Finally, we describe experimental results and analysis that validates our approaches.

\section{Approach}\label{sec:approach}

\textbf{Overview.} We build on the traditional two-stream ConvNet and explore the correlations between spatial and temporal streams by using two different proposed fusion frameworks.
\ZK{"We posit that rather than straightforwardly fusing features from each stream at the end, correlations and information across the streams and time can be leveraged to improve activity recognition." This sentence makes it seem like the concept is not well-established, but it is. We should tone down the part that's the same as other, and emphasize the story in introduction}
We specifically focus on two models that can be used to process temporal data: Temporal Segment LSTMs (TS-LSTM) which leverage recurrent networks and convolution over temporally-constructed feature matrices (Temporal-ConvNet)\ZK{Removed "Feature fusion from spatial and temporal streams are conducted separately using "}. Both methods achieve state of art performance, and the TS-LSTM surpasses existing methods. However, the details of the architectures matter significantly, and we show that the methods only work when used in particular ways. We perform significant experimentation to elucidate which design decisions are important. 
Figure \ref{fig:overview}  schematically illustrates our proposed methods.

\subsection{Two-stream ConvNets}
The two-stream ConvNet is constructed by two individual spatial-stream and temporal-stream ConvNets. The spatial-stream network takes RGB images as input, while the temporal-stream network takes stacked optical flow images as inputs. 
A great deal of literature has shown that using deeper ConvNets can improve overall performance for two-stream methods. In particular, the performance from VGG-16 \cite{simonyan2014very}, GoogLeNet \cite{szegedy2015going}, and BN-Inception \cite{icml2015_ioffe15} on both spatial and temporal streams are reported \cite{WangXWQLTV16,yue2015beyond,WangXWQLTV16}. 
Since ResNet-101 \cite{he2015deep} has demonstrated its capability in capturing still image features, 
we chose ResNet-101 as our ConvNet for both the spatial and temporal streams. We demonstrate that this can result in a strong baseline.
Feichtenhofer et al.~\cite{feichtenhofer2016convolutional} experiment with different fusion stages for the spatial and temporal ConvNets. Their results indicate that fusion can achieve the best performance using late fusions. Fusion conducted at early layers results in lower performance, though require less number of parameters. For this reason, we aim at exploring feature fusion using the last layer from both spatial-stream and temporal-stream ConvNets. In our framework, the two-stream ResNets serve as high-dimensional feature extractors. The output feature vectors at time step $t$ from the spatial-stream and temporal-stream ConvNets can be represented as $f_{t}^{S}\in \mathbb{R}^{n_{S}}$ and $f_{t}^{T}\in \mathbb{R}^{n_{T}}$, respectively. The input feature vector $x_t\in \mathbb{R}^{n_{S}+n_{T}}$ for our proposed temporal segment LSTM and Temporal-ConvNet is the concatenation of $f_{t}^{S}$ and $f_{t}^{T}$. In our case, $n_S$ and $n_T$ are both 2048.

\textbf{Spatial stream.}
Using a single RGB image for the spatial stream has been shown to achieve fairly good performance. 
Experimenting with the stacking of RGB difference is beyond the scope of this work, but can potentially improve performance~\cite{WangXWQLTV16}. The ResNet-101 spatial-stream ConvNet is pre-trained on ImageNet
and fine-tuned on RGB images extracted from UCF101 dataset with classification loss for predicting activities. 

\textbf{Temporal stream.}
Stacking 10 optical flow images for the temporal stream has been considered as a standard for two-stream ConvNets
\cite{simonyan2014two,feichtenhofer2016convolutional,yue2015beyond,WangXWQLTV16,WangSWVH16}. We follow the standard to show how each of our framework design and training practices can improve the classification accuracy. 
In particular, using a pre-trained network and fine-tuning has been confirmed to be extremely helpful despite differences in the data distributions between RGB and optical flow. 
We follow the same pre-train procedure shown by Wang et al.~\cite{WangXWQLTV16}.
The effectiveness of the pre-trained model on temporal-stream ConvNet is in Table \ref{table:temporal-comparison} in Section \ref{sec:implementation}.

\subsection{Temporal Segment LSTM (TS-LSTM)}
The variations between each of the image frames within a video may contain additional information that could be useful in determining the human action in the whole video. One of the most straightforward ways to incorporate and exploit sequences of inputs using neural networks is through a Recurrent Neural Network (RNN). RNNs can learn temporal dynamics from a sequence of inputs by mapping the inputs to hidden states, and from hidden states to outputs. The objective of using RNNs is to learn how the representations change over time for activities. However, several previous works have shown limited ability of directly using a ConvNet and RNN \cite{donahue2015long,yue2015beyond,pan2015jointly,abu2016youtube}. We therefore adapted temporal segments \cite{WangXWQLTV16} for use with RNNs and provide segmental consensus via temporal pooling and LSTM cells. In our work, the input of the RNN in each time step $t$ is a high-level feature representation $x_t\in \mathbb{R}^{4096}$. 









LSTM cells have been adapted for the action recognition problem, but so far have shown limited improvement~\cite{donahue2015long,yue2015beyond}. In particular, Ng et al.~\cite{yue2015beyond} 
use five stacked LSTM layers each with 512 memory cells. 
We argue that deeper LSTM layers do not necessarily help in achieving better action recognition performance, since the baseline two-stream ConvNet has already learned to become more robust in effectively representing frame-level images (see section \ref{sec:vanilla-lstm}).
\ZK{How do you define temporal variants? Be specific, and consistent since you use both variance and variants. Also, what is the distinction between stacked LSTM cells (which you say would perform equal) and deep LSTM networks (which you hypothesize would be worse?}
\Yao{Changed to use "variations". All I am trying to say is: one or two layers of LSTM works better than LSTM with many layers (see Table \ref{table:temporal-segment-lstm}). It's hard to justify the number of layers needed. It depends on whether if there are enough temporal variations for you to model.}
\ZK{Sure, but "variations" doesn't fully describe something. Variations across what? Spatial patterns across time? Temporal patterns themselves? Could be lots of things}
\Yao{I see. Changed all the other parts of paper mentioned "temporal variations" to something more specific.}

\begin{figure}[!tbp]
\begin{center}
   \includegraphics[width=.75\linewidth]{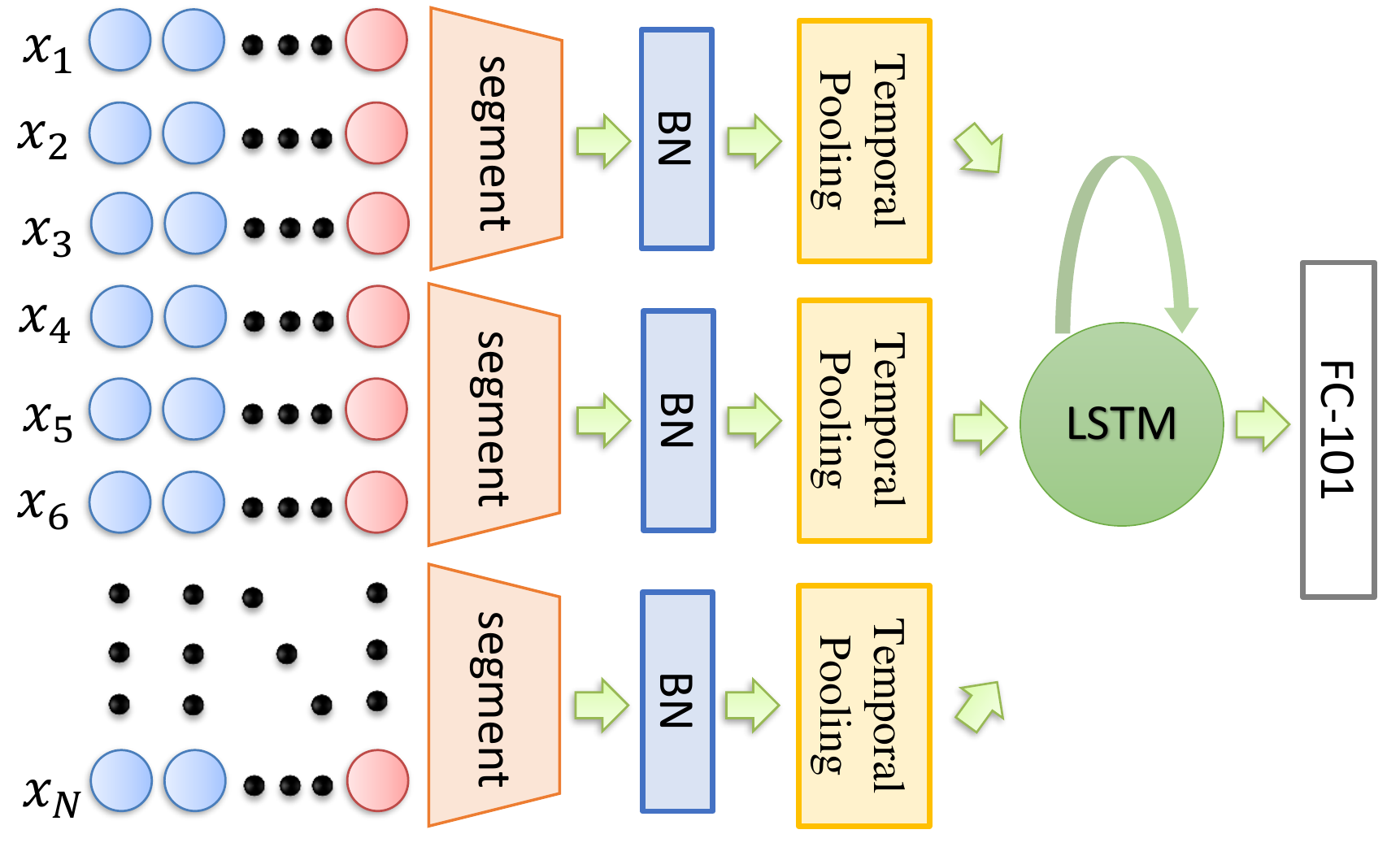}
\end{center}
   \caption{\textbf{Temporal Segment LSTM (TS-LSTM)} first divides the feature matrix into several temporal segments. Each temporal segment is then pooled via mean or max pooling layers, and their outputs are fed into the LSTM layer sequentially. 
   }
\label{fig:ts-lstm}
\end{figure}

\textbf{Temporal segments and pooling.}
In addressing the limited improvement of using LSTM, we follow the intuition of the temporal segment by Wang et al.~\cite{WangXWQLTV16} and divide the sampled video frames into several segments. A temporal pooling layer is applied to extract distinguishing features from each of the segments, and an LSTM layer is used to extract the embedded features from all segments. The proposed TS-LSTM is shown in Figure \ref{fig:ts-lstm}. 

First, the concatenated spatial and temporal features will be pooled through temporal pooling layers, e.g. mean or max pooling. In practice, the temporal mean pooling performed similarly with max pooling. We use max pooling in our experiments. The temporal pooling layer takes the feature vectors concatenated 
from spatial and temporal streams and extracts distinguishing feature elements. The recurrent LSTM module then learns the final embedded features for the entire video. The proposed TS-LSTM module essentially serves as a mechanism that learns the non-linear feature combination and its segmental representation over time. We discuss implications of the success of TS-LSTMs in the sections describing experimental results.
\subsection{Temporal-ConvNet}
\textbf{Spatiotemporal correlations.} One of the main drawbacks of 
baseline two-stream ConvNets is that
the network only learns the spatial correlation within a single frame instead of leveraging the temporal relation across different frames. To address this issue, \cite{sun2015human, feichtenhofer2016convolutional} adopted the concept of 3D kernels to exploit the pixel-wise correlation between spatial and temporal feature maps. However, both approaches applied convolution kernels of one scale to extract the temporal features with fixed temporal receptive fields, and they did so on the full feature maps which results in more than 40 times the parameters compared to using feature vectors (see supplementary material).\Steve{I temporarily remove the description related to dimensionality. \cite{sun2015human} mapped 51*51*T (frame number)*512 to 4096, and \cite{feichtenhofer2016convolutional} mapped 56*56*T*512 to 4096. Both linear layers require lots of parameters. In our case, we extract the 2048-dimension feature vectors from ResNet and concatenate the two streams to form a 4096-dimension vector. Our Temporal-ConvNet reduces the dimension from 4096*T to 4096 with TCL \ZK{You don't define TCL until later} and then maps to 1024, so we only need 4M parameters. If we extract the feature maps from the previous layer with the dimension 7x7x4096 as our inputs, we will need 49 times more parameters in our architecture. To reduce the dimension, we need to change the architecture (at least using 3D kernels), so the computation will be much higher. The architecture in  \cite{feichtenhofer2016convolutional} needs more than 90M parameters. \cite{sun2015human} didn't mention the parameter number. But I am not sure if I should include all the details in the paper, or how I should describe this.}\ZK{Space is limited but if it is backed up it is good to mention. Perhaps mention it and put some statistics/parameter numbers in the supplementary.}\Steve{I will put all the description in the comments to supplementary.} In contrast to these approaches, we focus on designing a more efficient architecture with multiple convolution layers to explore the temporal information only using feature vectors.

In addition to using RNNs to learn the temporal dynamics, we adapt the ConvNet architecture on feature matrices $\textit{\textbf{x}} = \{x_1, x_2, ..., x_t, ..., x_N\}\in \mathbb{R}^{4096\times N}$, where $N$ is the number of sampled frames in one video. Different from natural images, the elements in each $x_t$ have little spatial dependency, but have temporal correlation across different $x_t$.
With the ConvNet architecture, we explore the temporal correlation of the feature elements, and then distinguish different categories of actions. 

\begin{figure}[!htbp]
\begin{center}
   \includegraphics[width=1\linewidth]{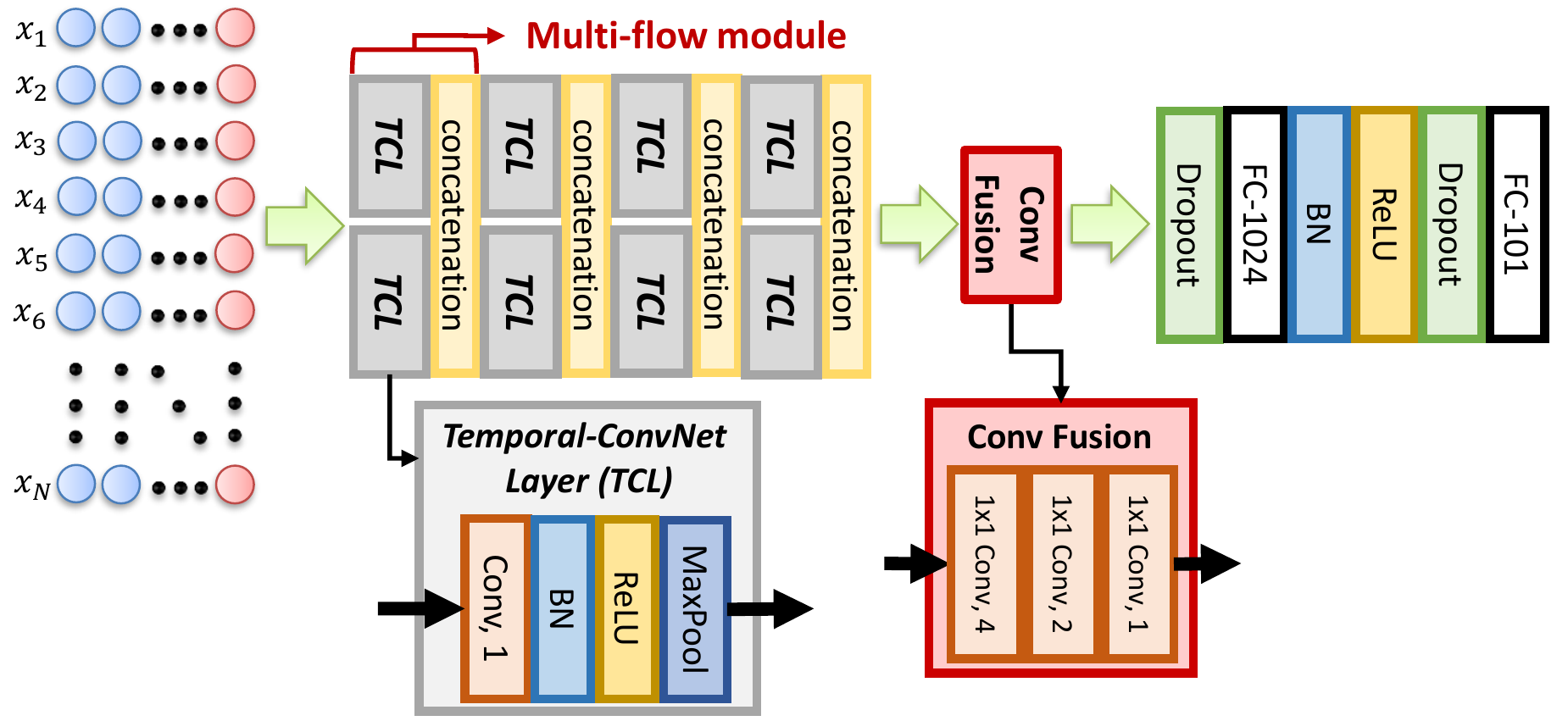}
\end{center}
   \caption{\textbf{Overall Temporal-Inception architecture.} 
   The input of Temporal-Inception is a 2D matrix composed of feature vectors across different time steps. In each multi-flow module, there are two \textit{TCL}s with different convolution kernels. Each multi-flow module reduces the temporal dimension by half. Therefore, with multiple \textit{TCL}s and two fully-connected layers, the input feature matrices are mapped to the class prediction.
   } 
\label{fig:tem-conv}
\end{figure}

The overall architecture of the Temporal-ConvNet is composed of multiple \textit{Temporal-ConvNet layers (TCLs)}, as shown in Figure \ref{fig:tem-conv}, and can be formulated as follows: 



$TemConv(\textit{\textbf{x}})$ 

$= H(G(F(F(F(F(\textit{\textbf{x}},\mathbf{W_{1}}),\mathbf{W_{2}}),\mathbf{W_{3}}),\mathbf{W_{4}})))$

where
$F(\textit{\textbf{x}},\mathbf{W_{i}}) = (TCL_1(\textit{\textbf{x}};\mathbf{W_{i_{1}}}),TCL_2(\textit{\textbf{x}};\mathbf{W_{i_{2}}}))$

$G$ maps the output of the final \textit{TCL} to a 1024-dimension vector. $H$ is the Softmax function. $\mathbf{W_{i_{j}}}$ denotes the parameter set used for \textit{TCL}, where i denotes layer number and j represents the index of \textit{TCL} in a multi-flow module. 

\textbf{Temporal-Inception model.} After obtaining the feature matrix $\textit{\textbf{x}}$, we apply 1D kernels to specifically encode temporal information in different scales and reduce the temporal dimension because the spatial information is already encoded from the two-stream ConvNet. Applying convolution along the spatial direction may alter the learned spatial characteristics. In addition, we adapt the inception module \cite{szegedy2015going} into our architecture and note it as the multi-flow module, which consists of different convolution kernel sizes, as shown in Figure~\ref{fig:tem-conv}. Multiple multi-flow modules are used to hierarchically encode temporal information and reduce the temporal dimension. However, the filter dimension gradually increases because of the concatenation in each multi-flow module. To avoid potential overfitting issues, we convolve the concatenated feature vectors with a set of filters $\mathbf{f_{i}} \in \mathbb{R}^{1\times 1\times D_{i1}\times D_{i2}}, D_{i2} < D_{i1},$ to reduce the filter dimension. 
We note this architecture as \textit{Temporal-Inception} in this paper. 
The rationale behind \textit{Temporal-Inception} is that different types of action have different temporal characteristics, and different kernels in different layers essentially search for different actions by exploiting different receptive fields to encode the temporal characteristics.

The current state-of-the-art method from Wang et al.~\cite{WangXWQLTV16} and our Temporal-Inception both explore the temporal information but with different perspectives  and can be complementary. \cite{WangXWQLTV16} focuses on designing novel and effective sampling approaches for temporal information exploration while we focus on designing the architecture to extract the temporal convolutional features given the sampled frames. 


\subsection{Implementation} \label{sec:implementation}
\subsubsection{Training and testing practice}
\textbf{Two-stream inputs.}
We use both RGB and optical flow images as inputs to two-stream ConvNets.
For generating the optical flow images, literature have different choices for optical flow algorithms. Although most of the works used either Brox \cite{brox2004high} or TV-L1 \cite{zach2007duality}, within each of the different optical flow algorithms there are still some variations in how the optical flow images are thresholded and normalized. We summarize the prediction accuracy of different optical flow methods from recent works in Table \ref{table:temporal-comparison}. 
Note that we thresholded the absolute value of motion magnitude to 20 and rescale the horizontal and vertical components of the optical flow to the range [0, 255] for TV-L1. 
From Table \ref{table:temporal-comparison}, we can conclude that both Brox and TV-L1 can achieve state-of-the-art performance, but from our experiments TV-L1 is slightly better than Brox. Thus, unless specified we use TV-L1 as input for the temporal-stream ConvNet. 

\begin{table}[!tbp]
\small
\centering
\caption{Optical flow algorithms and Temporal-stream ConvNet performance comparison.}
\label{table:temporal-comparison}

\resizebox{\columnwidth}{!}{
\begin{tabular}{c|cccc}
& Optical flow  & ConvNet      & Fine-tune & Accuracy \\ \hline
Two-stream~\cite{simonyan2014two} & Brox & CNN M 2048   & N         & 81.0     \\ \hline
\multirow{2}{*}{\begin{tabular}[c]{@{}c@{}}Convolutional\\ Two-stream~\cite{feichtenhofer2016convolutional}\end{tabular}} & \multirow{2}{*}{\begin{tabular}[c]{@{}c@{}}Brox\\ {[}-20,20{]}\end{tabular}} & VGG-M        & Y         & 82.3     \\ & & VGG-16       & Y         & 86.3     \\ \hline
\multirow{2}{*}{TSN~\cite{WangXWQLTV16}} & \multirow{2}{*}{TV-L1} & VGG-16       & Y         & 85.7     \\ 
& & BN-Inception & Y         & 87.2     \\ \hline
SR-CNNs~\cite{WangSWVH16} & TV-L1 & VGG-16       & Y         & 85.3     \\ \hline
Ours & \begin{tabular}[c]{@{}c@{}}TV-L1\\ {[}-20,20{]}\end{tabular}                 & ResNet-101   & Y         & 86.2     \\ \hline
Ours & Brox                                                                         & ResNet-101   & Y         & 84.9    
\end{tabular}
}
\end{table}

\textbf{Hyper-parameter optimization. }
The learning rate of the spatial-stream ConvNet is initially set to \num{5e-6}, and divided by 10 when the accuracy is saturated. The weight decay is set to be \num{1e-4}, and momentum is 0.9. On the other hand, the learning rate of the temporal-stream ConvNet is initially set to \num{5e-3}, and divided by 10 when the accuracy is saturated. The weight decay and momentum are the same as the spatial-stream ConvNet. The batch sizes for both ConvNets are 64.
Both Temporal Segment LSTM and Temporal-ConvNets are trained with ADAM optimizer. The learning rate is set to \num{5e-5} for training Temporal Segment LSTM. For Temporal-ConvNets, we use \num{1e-4} for learning rate and \num{1e-1} for weight decay.

\textbf{Data augmentation.}
Data augmentation has been very helpful especially when the training data are limited. 
During training, a sub-image with size 256 x 256 is first randomly cropped using a smaller image region (between 0.08 to 1 of the original image area). 
Second, the cropped image was randomly scaled between 3/4 and 4/3 of its size. Finally, the cropped and scaled image will be scaled again to 224 x 224. The same data augmentation is applied when training both spatial- and temporal-stream ConvNets. Note that we use additional color jittering for the spatial-stream ConvNet, but not temporal-stream ConvNet. Wang et al.~\cite{WangXWQLTV16} argued that a corner cropping strategy, which only crops 4 corners and center of the image, can prevent overfitting. However, in our implementation, the corner cropping strategy actually makes the ConvNet converge faster and leads to overfitting.

\textbf{Testing.}
We followed the work from Simonyan and Zisserman~\cite{simonyan2014two} to use 25 frames for testing. Each of the 25 frames is sampled equally across each of the videos. 
During testing, many works averaged predictions from the RNN on all 25 frames. We did not average the prediction because it will be biased towards the average representation of the video and neglect the temporal dynamics learned by LSTM cells. However, without temporal segments, since LSTM cells fail to learn the temporal dynamics, it will benefit from averaging the predictions. Yet, the final prediction accuracy is significantly worse than temporal segments LSTM without averaging prediction. 



\section{Evaluation}\label{sec:evaluation}
We adapt our proposed model to two of the most widely used action recognition benchmarks: UCF101~\cite{soomro2012UCF101} and HMDB51~\cite{kuehne2011hmdb}. 
We used the first split of UCF101 for validating our proposed models. 
We used the same parameters and architectures obtained from UCF101 split 1 directly for the other two splits. The model pre-trained on UCF101 is fine-tuned for the HMDB51 dataset. 

\subsection{Baseline Two-stream ConvNets.}
Table \ref{table:baseline-comparison} shows our experiment results for the spatial-stream, temporal-streams, and two-stream on three different splits in the UCF101 and HMDB51 datasets. Our performance on the two-stream model is obtained by taking the mean values of the prediction probabilities from both spatial and temporal-stream ConvNets. Our two proposed methods leverage the baseline two-stream ConvNet and show significant improvement by modeling temporal dynamics. 
For comparison of our baseline method with others, please refer to supplementary material. 


\begin{table}[!htbp]
\centering
\small
\caption{Performance from spatial and temporal-stream ConvNets, and two-stream ConvNet on three different splits of the UCF101 and HMDB51 datasets.}
\label{table:baseline-comparison}
\begin{tabular}{ccccc}
\multicolumn{1}{c|}{}       & Spilit 1 & Split 2 & \multicolumn{1}{c|}{Split 3} & Mean \\ \hline
\multicolumn{5}{c}{\textbf{Spatial-stream}}  \\ \hline
\multicolumn{1}{c|}{UCF101} & 86.1     & 83.6    & \multicolumn{1}{c|}{85.3}    & 85.0 \\
\multicolumn{1}{c|}{HMDB51} & 51.9     & 49.7    & \multicolumn{1}{c|}{49.7}    & 50.4 \\ \hline
\multicolumn{5}{c}{\textbf{Temporal-stream}} \\ \hline
\multicolumn{1}{c|}{UCF101} & 86.2     & 87.0    & \multicolumn{1}{c|}{88.4}    & 87.2 \\
\multicolumn{1}{c|}{HMDB51} & 60.3     & 59.0    & \multicolumn{1}{c|}{60.0}    & 59.7 \\ \hline
\multicolumn{5}{c}{\textbf{Two-stream}}      \\ \hline
\multicolumn{1}{c|}{UCF101} & 92.6     & 92.2    & \multicolumn{1}{c|}{92.9}    & 92.6 \\
\multicolumn{1}{c|}{HMDB51} & 66.4     & 64.8    & \multicolumn{1}{c|}{92.6}    & 64.6
\end{tabular}
\end{table}

\subsection{Performance of TS-LSTM}

In the following section, we discuss various experiments and individual performance associated with different architectural designs for TS-LSTM. We thoroughly explore different network designs and conclude that: (i) using temporal segments perform better than no segments (ii) adding a FC layer before temporal pooling layer can \ZK{What is the architecture without LSTM? Which FC layer? Is it from the two-stream baseline?}
\Yao{without LSTM: only temporal segments and max pooling. But because the we concatnate all segments, your dimension increase to 3 x 4096, which leads to overfitting. Thus, adding a FC before the max pooling can reduce the feature dimension and prevent overfitting. Changed the desciption, so it's less confusing.}learn the spatial and temporal feature integration and reduce the feature dimension for each segment. (iii)  LSTMs effectively replace the needs for the first FC layer and learns the spatial and temporal feature integration while reducing the feature dimension. 
(iv) deep LSTM layers do not necessarily help and often lead to overfitting 
These experiments lead us to conclude that despite their theoretical ability to extract temporal patterns over many scales and lengths, in this case their inputs must be pre-segmented, demonstrating a limitation of our current usage of LSTMs. The summarization of our experiments on TS-LSTM is shown in Table \ref{table:temporal-segment-lstm}. 

\textbf{Temporal segments and pooling.}
Temporal segments have been shown to be useful in end-to-end frameworks \cite{WangXWQLTV16}. In our experiments, we demonstrate that even with pre-saved feature vectors extracted from equally sampled images in the videos, temporal segments can help in improving the classification accuracy. The difference of using three or five segments is statistically insignificant and may highly depend on the types of action performed in the video. Temporal mean pooling performed similarly with max pooling. We use max pooling for the rest of experiments. 


\textbf{Spatial and temporal feature integration.}
Since the features from each segment are concatenated together, the model suffers from overfitting when the number of segments increases, as can be seen from Table \ref{table:temporal-segment-lstm} with five segments. By adding an FC layer before the temporal max pooling layer, we can effectively control the dimensionality of the embedded features, exploit the spatial and temporal correlation, and prevent overfitting.

\textbf{Vanilla LSTM.} \label{sec:vanilla-lstm}
LSTM cells have the ability to model temporal dynamics, but only shown limited improvement from previous works. Our experimental results shown that there is only a 0.2\% improvement over two-stream ConvNet, which is consistent with \cite{yue2015beyond} (88.0 to 88.6\%) and \cite{donahue2015long} (69.0 to 71.1\% on RGB, 72.2 to 77.0\% on flow), and \cite{abu2016youtube} (63.3 to 64.5\%). The performance gained from LSTM cells decreases when the baseline two-stream ConvNet is stronger\ZK{Remofed: more robust in representing dynamic temporal information. Kind of a conjecture, especially since it doesn't really represent temporal correlations}. \Yao{agree. Changed the other part of paper to make the claim consistent.} Thus, we observed only 0.2\% improvement when using a baseline two-stream ConvNet achieving 92.6\% accuracy, as opposed to 2-5\% improvement from a baseline of 70\% accuracy ~\cite{donahue2015long}. Note that using vanilla LSTM performed only similar to naive temporal max pooling.

\textbf{TS-LSTM.}
By combining the temporal segments and LSTM cells, we can leverage the temporal dynamics across each temporal segment, and significantly boost the prediction accuracy. 
This finding suggests that carefully re-thinking and understanding how LSTMs model temporal information is necessary\ZK{This isn't for this paper, but I'm wondering if skip connections wouldn't be useful here, to allow the LSTM to better skip or pool information across sub-segments. Probably some text literature exploration would be helpful, e.g. https://arxiv.org/abs/1610.03167?context=cs}. \Yao{Agree. I will explore this direction after this paper.} 
Our experimental results indicate that using deeper LSTM layers is prone to overfitting on the UCF101 dataset. This is probably because our features generated from spatial and temporal ConvNets were fine-tuned to identify video classes at the frame level. The dynamics of feature representations over time is not as complicated as other sequential data, e.g. speech and text. Thus, increasing the number of stacked LSTM layers tends to overfit the data in UCF101. Our experiments on HMDB51 also confirm this hypothesis. The prediction accuracy on the HMDB51 dataset using two-layer LSTM increased from 68.7\% to 69.0\% over single LSTM layer, since the baseline model has not yet learned to robustly identify video classes at the frame level.

\begin{table}[!tbp]
\small
\centering
\caption{Comparison of each component in Temporal Segment LSTM on UCF101 split 1. TS: number of temporal segments. Max: temporal max pooling layer. 512: dimension of LSTM cell. For a more complete version, please refer to supplementary material. \Yao{remove 5 segments to save space.}} 
\label{table:temporal-segment-lstm}
\begin{tabular}{ccccccc}
TS    & BN    & FC      & Temporal Pooling          & BN    & \multicolumn{1}{c}{FC}     & Acc      \\ \hline
\multicolumn{7}{c}{\textbf{LSTM}} \\ \hline
1           & BN    &     & 512                     & BN    & \multicolumn{1}{c|}{101}    & 92.8    \\ 
1           & BN    &     & (512,512)               & BN    & \multicolumn{1}{c|}{101}    & 92.5    \\ 
1           & BN    &     & (512,512,512)           & BN    & \multicolumn{1}{c|}{101}    & 92.1    \\ \hline
\multicolumn{7}{c}{\textbf{Temporal segment \& Batch normalization}}                                \\ \hline
1           & BN    &         & Max                 & BN    & \multicolumn{1}{c|}{101}    & 92.8    \\
3           & BN    &         & Max                 & BN    & \multicolumn{1}{c|}{101}    & 93.4    \\
\multicolumn{7}{c}{\textbf{Feature integration \& dimension reduction}} \\ \hline
3           & BN    & 512     & Max                 & BN    & \multicolumn{1}{c|}{101}    & 93.9    \\
\multicolumn{7}{c}{\textbf{Temporal Segment LSTM}}                                                  \\ \hline
3           & BN    &     & Max + 512               & BN    & \multicolumn{1}{c|}{101}    & \textbf{94.3}    \\
3           & BN    &     & Max + (512,512)         & BN    & \multicolumn{1}{c|}{101}    & \textbf{94.2}    \\
3           & BN    &     & Max + (512,512,512)     & BN    & \multicolumn{1}{c|}{101}    & 93.9    \\
\end{tabular}
\end{table}

\subsection{Performance of Temporal-ConvNet}
In this Section, we discuss different factors for designing the architecture of the Temporal-ConvNet. We conclude that: (i) applying multiple \textit{TCL}s performs better than using single or double \textit{TCL}s. (ii) concatenating the outputs of each multi-flow module is the better way to fuse different flows. (iii) with proper fusion methods, the multi-flow architecture has better capability to explore 
the temporal information than the single-flow architecture. (iv) adding batch normalization and dropout layers can further improve the performance. The summarization of our experiments on Temporal-ConvNet is shown in Table \ref{table:temporal-convnet}. Based on these experiments, we can conclude that convolution across time can effectively extract temporal patterns, but as with other applications the specific architecture is crucial and lessons learned from other architectures and tasks can inform successful designs. 

\textbf{Temporal-ConvNet layer.} 
One of the most important components in the Temporal-ConvNet is the \textit{TCL}, and the convolutional kernel size is the most critical part since it directly affects how the network learns the temporal correlation. The different kernel sizes essentially correspond to actions with different temporal duration and period. We set the temporal convolutional kernel size to 5 for our experiments. On the other hand, the number of \textit{TCL}s also plays an important role, because the \textit{TCL}s are used to gradually reduce the dimension in the temporal direction, i.e. we map the feature matrices to a feature vector. 
Applying only single or double \textit{TCL}s will still leave the temporal direction to have a high dimensionality, and thus result in larger numbers of parameters and cause overfitting. 
Table \ref{table:temporal-convnet} shows the results from different numbers of \textit{TCL}s.

\begin{figure}[!htbp]  
    \centering
    \begin{tabular}[c]{ccc}
    \begin{subfigure}[c]{0.1\textwidth}
        \includegraphics[width=\textwidth]{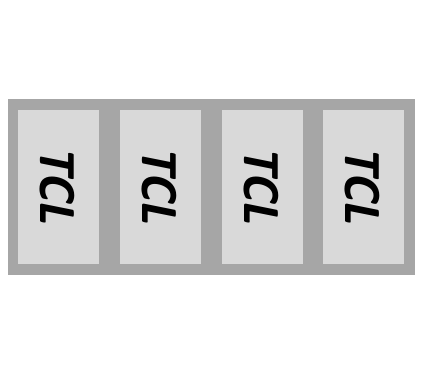}
        \caption{Temporal-VGG}
        \label{fig:conv-vgg}
    \end{subfigure}&
    \begin{subfigure}[c]{0.12\textwidth}
        \includegraphics[width=\textwidth]{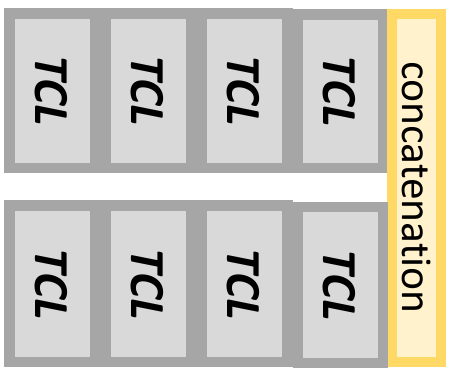}
        \caption{Multi-flow Temporal-VGG}
        \label{fig:multi-flow_conv-vgg}
    \end{subfigure}&
    \begin{subfigure}[c]{0.18\textwidth}
        \includegraphics[width=\textwidth]{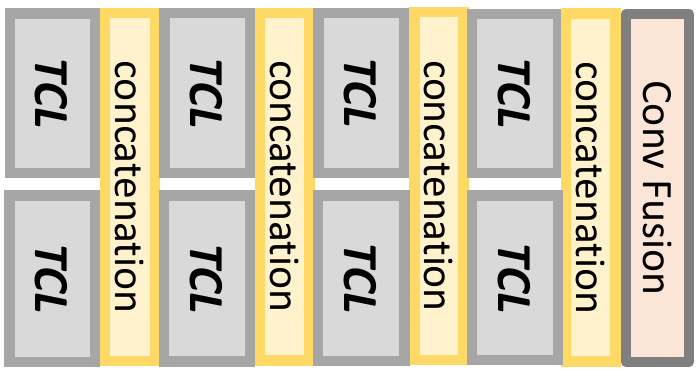}
        \caption{Temporal-Inception}
        \label{fig:conv-inception}
    \end{subfigure}
    \end{tabular}
    \caption{\textbf{Comparison of three different architecture of Temporal-ConvNet.} \textit{Temporal-Inception} has the best performance.}
    \label{fig:tem-conv_comparison}
\end{figure}

\textbf{Multi-flow architecture}
There are two main questions in optimizing the multi-flow architecture: 1) How to combine multiple flows? 2) How many flows should we have? 
For the first question, we propose two approaches. One is our \textit{Temporal-Inception}, and the other one is the multi-flow version of \textit{Temporal-VGG}. The difference between these two approaches is where we place the concatenation layers, as shown in Figure \ref{fig:tem-conv_comparison}. 
Regarding the second question, increasing the flow number provides a better capability to describe actions with different temporal scales, but it also greatly increases the chance to overfit the data. In addition, with the multi-flow architecture, the size of the temporal convolutional kernel is also important. From our experiments, kernel sizes of 5 and 7 achieved the best prediction accuracy. 

The illustration of \textit{Temporal-VGG}, \textit{Multi-flow Temporal-VGG}, and \textit{Temporal-Inception} are shown in Figure \ref{fig:tem-conv_comparison}. The prediction accuracy on UCF-101 split 1 is shown in Table \ref{table:temporal-convnet}. The \textit{Temporal-Inception} has better performance than the multi-flow version of \textit{Temporal-VGG}, since \textit{Temporal-Inception}  effectively explores and combines temporal information obtained through various temporal receptive fields.

\textbf{Batch normalization and dropout} \Steve{The importance of this part is minor. We can consider to take away this part.}
To overcome the overfitting and internal covariate shift issues, we add a batch normalization layer right before each ReLU layer, and add dropout layers before and after the FC-1024 layer. We use \textit{Temporal-Inception} to demonstrate how batch normalization and dropout improve the performance.
\Yao{Steve, your description here is comparing Temporal-Inception without BN and dropout with Temporal-VGG with BN and dropout. That's not a fair comparison. And, I don't think it's important to mention these too much. I changed this paragraph.}

\begin{table}[!htbp]
\small
\centering
\caption{Performance of Temporal-ConvNet on UCF101 split 1. 1L: single \textit{TCL}. 2L: double \textit{TCL}. BN: Batch Normalization. FC: fully-connected layer. In the column ``Architecture", ``T" is denoted as one \textit{TCL}. \{\} denote as the stacked architecture. () denotes as the wide (parallel) architecture. The illustration of the last three methods are shown in Figure \ref{fig:tem-conv_comparison}.}
\label{table:temporal-convnet}
\resizebox{0.95\columnwidth}{!}{
\begin{tabular}{ccccc}
Architecture            & BN    & Dropout & \multicolumn{1}{c}{FC}      & Acc   \\ \hline
\multicolumn{5}{c}{\textbf{1L}} \\ \hline
T                   & BN    & Dropout & \multicolumn{1}{c|}{1024}   & 93.6    \\  
(T,T)               & BN    & Dropout & \multicolumn{1}{c|}{1024}   & 92.6    \\  \hline
\multicolumn{5}{c}{\textbf{2L}} \\ \hline
\{T,T\}                 & BN    & Dropout & \multicolumn{1}{c|}{1024}     & 93.1    \\  
\{(T,T),(T,T)\}     & BN    & Dropout & \multicolumn{1}{c|}{1024}     & 92.7    \\  \hline
\multicolumn{5}{c}{\textbf{Temporal-VGG}} \\ \hline
\{T,T,T,T\}               & BN    & Dropout& \multicolumn{1}{c|}{1024}    & 94.0    \\  \hline
\multicolumn{5}{c}{\textbf{Multi-flow Temporal-VGG}} \\ \hline
(\{T,T,T,T\},\{T,T,T,T\})     & BN    & Dropout & \multicolumn{1}{c|}{1024}   & 93.4    \\  \hline
\multicolumn{5}{c}{\textbf{Temporal-Inception}} \\ \hline
\multicolumn{1}{c|}{} &     &         & \multicolumn{1}{c|}{1024}   & 93.3              \\
\multicolumn{1}{c|}{} &    & Dropout  & \multicolumn{1}{c|}{1024}   & 93.4    \\       
\multicolumn{1}{c|}{\{(T,T),(T,T),(T,T),(T,T)\}}    & BN    &     & \multicolumn{1}{c|}{1024}   & 93.4    \\
\multicolumn{1}{c|}{} & BN & Dropout  & \multicolumn{1}{c|}{2048}   & 93.7    \\       
\multicolumn{1}{c|}{} & BN & Dropout  & \multicolumn{1}{c|}{1024}   & \textbf{94.2}   
\end{tabular}
}
\end{table}

\subsection{Final Performance}

The results from the proposed Temporal Segment LSTM and Temporal-Inception on both UCF101 and HMDB51 are shown in Table \ref{tabel:state-of-the-art}. 
While TSN \cite{WangXWQLTV16} achieved significant progress on human action recognition, it requires significant temporal augmentation and it is unclear how such augmentation and temporal segments each contributed to the results. \ZK{Removed: help in making the model generalize. This is kind of a generic claim and its basis is unclear (I guess besides a literal interpretation of better generalization implying better overall accuracy.}\Yao{Agree.}Our proposed methods explore temporal segments and demonstrate that, without tedious randomly sampled snippets from video in each training step, a simple temporal pooling layer and LSTM cells trained on a fixed sampled video can achieve better accuracy on both UCF101 and HMDB51.


\begin{table}[]
\small
\centering
\caption{State-of-the-art action recognition comparison on the UCF101~\cite{soomro2012UCF101} and HMDB51~\cite{kuehne2011hmdb} datasets.}
\label{tabel:state-of-the-art}
\begin{tabular}{ccc}
Methods                                                        & UCF101 & HMDB51 \\ \hline
F$_{st}$CN \cite{sun2015human}                                 & 88.1 & 59.1 \\
Two-stream \cite{simonyan2014two}                              & 88.0 & 59.4 \\
LSTM \cite{yue2015beyond}                                      & 88.6 & -    \\
Transformation \cite{Wang_2016_CVPR}                           & 92.4 & 63.4 \\
Convolutional Two-stream \cite{feichtenhofer2016convolutional} & 92.5 & 65.4 \\
SR-CNN \cite{WangSWVH16}                                       & 92.6 & -    \\
Key volume \cite{Zhu_2016_CVPR}                                & 93.1 & 67.2 \\
ST-ResNet \cite{feichtenhofer2016spatiotemporal}               & 93.4 & -    \\
TSN (2 modalities) \cite{WangXWQLTV16}                         & 94.0 & 68.5 \\ \hline
\textbf{TS-LSTM}                                               & \textbf{94.1} & \textbf{69.0} \\ 
\textbf{Temporal-Inception}                                      & \textbf{93.9} & \textbf{67.5} \\ \hline
\end{tabular}
\end{table}




\textbf{Modeling temporal dynamics.}\label{sec:temporal-dynamics}
To further validate that our methods can model the temporal dynamics, we train both TS-LSTM and Temporal-Inception using only a maximum of the first 10 seconds of videos, e.g. 250 frames per video. Note that the number of frames in the UCF101 dataset ranges from 30 to 1700. About 21\% of videos are longer than 10 seconds. Our experiments show that by only using the first 10 seconds of the video, the baseline two-stream ConvNets achieve 92.9\% accuracy which is slightly better than when using the full length of the videos.
On the other hand, when the proposed methods see more frames of each video, both of the methods achieved better prediction accuracy (TS-LSTM: 93.7 to 94.1\%; Temporal-Inception: 93.2 to 93.9\%). This verifies that our proposed methods can successfully leverage the temporal information. 
\section{Conclusion \& Discussion}
Two-stream ConvNets have been widely used in video understanding, especially for human action recognition. Recently, several works have explored various methods to exploit spatiotemporal information. However, these works have tried individual methods with little analysis of whether and how they can successfully model dynamic temporal information, and often multiple differences obfuscate the exact cause for better performance. 
In this paper, we thoroughly explored two methods to model dynamic temporal information: Temporal Segment LSTM and Temporal-ConvNet. We showed that naive temporal max pooling performed similar to the vanilla LSTM. By integrating temporal segments and LSTM, the proposed method achieved state-of-the-art accuracy on both UCF101 and HMDB51. Our proposed Temporal-ConvNet performs convolutions on temporally-constructed feature vectors to learn global video-level representations. We further investigated different VGG- and Inception-style Temporal-ConvNets, and demonstrated that the proposed Temporal-Inception can achieve state-of-the-art performance using only high-level feature vector representations equally sampled from each of the videos. Combined, we show that both RNNs and convolutions across time are able to model temporal dynamics, but that care must be given to account for strengths and weaknesses in each approach. 
Our findings using temporal segments with LSTMs suggests that there may be low-hanging fruit in carefully re-thinking and understanding how LSTMs model temporal information. On the other hand, it is clear that convolutional networks can exploit information across time as well, and that design choices used for spatial tasks can transfer to temporal tasks. In the future, we plan to pursue these directions on bigger datasets as well as investigate how to better regularize LSTMs. \Yao{I removed "when data is scarce", because YouTube8M is a big dataset but still suffer from the same issue.} 

\appendix
\section{Appendix}

\subsection{Two-stream ConvNets Comparison on UCF101}
A great deal of literature has shown that using deeper ConvNets can improve overall performance for two-stream methods. In particular, the performance of VGG-16 \cite{simonyan2014very}, GoogLeNet \cite{szegedy2015going}, and BN-Inception \cite{icml2015_ioffe15} on both spatial and temporal streams are reported \cite{WangXWQLTV16,yue2015beyond}. Table \ref{table:twostream} compares the baseline performance using different ConvNets for training spatial- and temporal-stream ConvNet. We demonstrate a strong baseline performance using ResNet-101~\cite{he2015deep}. We expect to observe better performance by training with Inception-v3 or v4 models~\cite{szegedy2016inception,szegedy2015inceptionv2}. 
\begin{table*}[!htbp]
\small
\centering
\caption{Two-stream ConvNet comparison on the UCF101 dataset.}
\label{table:twostream}
\begin{tabular}{c|c|c|c}
             & Spatial-stream ConvNet & Temporal-stream ConvNet & Two-stream ConvNet \\ \hline
GoogLeNet \cite{simonyan2014two}    & 77.1                 & 83.9                  & 89.0             \\ \hline
VGG-16 \cite{wang2015towards}       & 78.4                 & 87.0                  & 91.4             \\ \hline
BN-Inception \cite{WangXWQLTV16}    & 84.5                 & 87.2                  & 92.0             \\ \hline
ResNet-101                          & 85.0                 & 87.2                  & \textbf{92.6}            
\end{tabular}
\end{table*}

\subsection{Complete Experimental Results for TS-LSTM}
We demonstrate the effectiveness of the proposed TS-LSTM and its comparison with various networks and dimensions we used in developing TS-LSTM. We show how vanilla LSTM can be properly trained and largely benefited by using batch normalization, but the vanilla LSTM still overfits the training samples and sometimes results in lower accuracy than the baseline two-stream method (92.6\%). A naive temporal pooling method can perform similarly with vanilla LSTM. We can further increase the accuracy by integrating temporal segments with LSTMs. We also show how a different number of temporal segments affects the prediction accuracy when using different dimension and the depth of the LSTM cells. Using three or five temporal segments results in similar performances in our experiments on UCF101 split 1. It is worth mentioning that the number of temporal segments might depend on the type of video classification problem we are trying to solve. Different types of video or actions may benefit from employing more temporal segments. 

\begin{figure}[!htbp]
    \centering
    \includegraphics[width=0.5\textwidth]{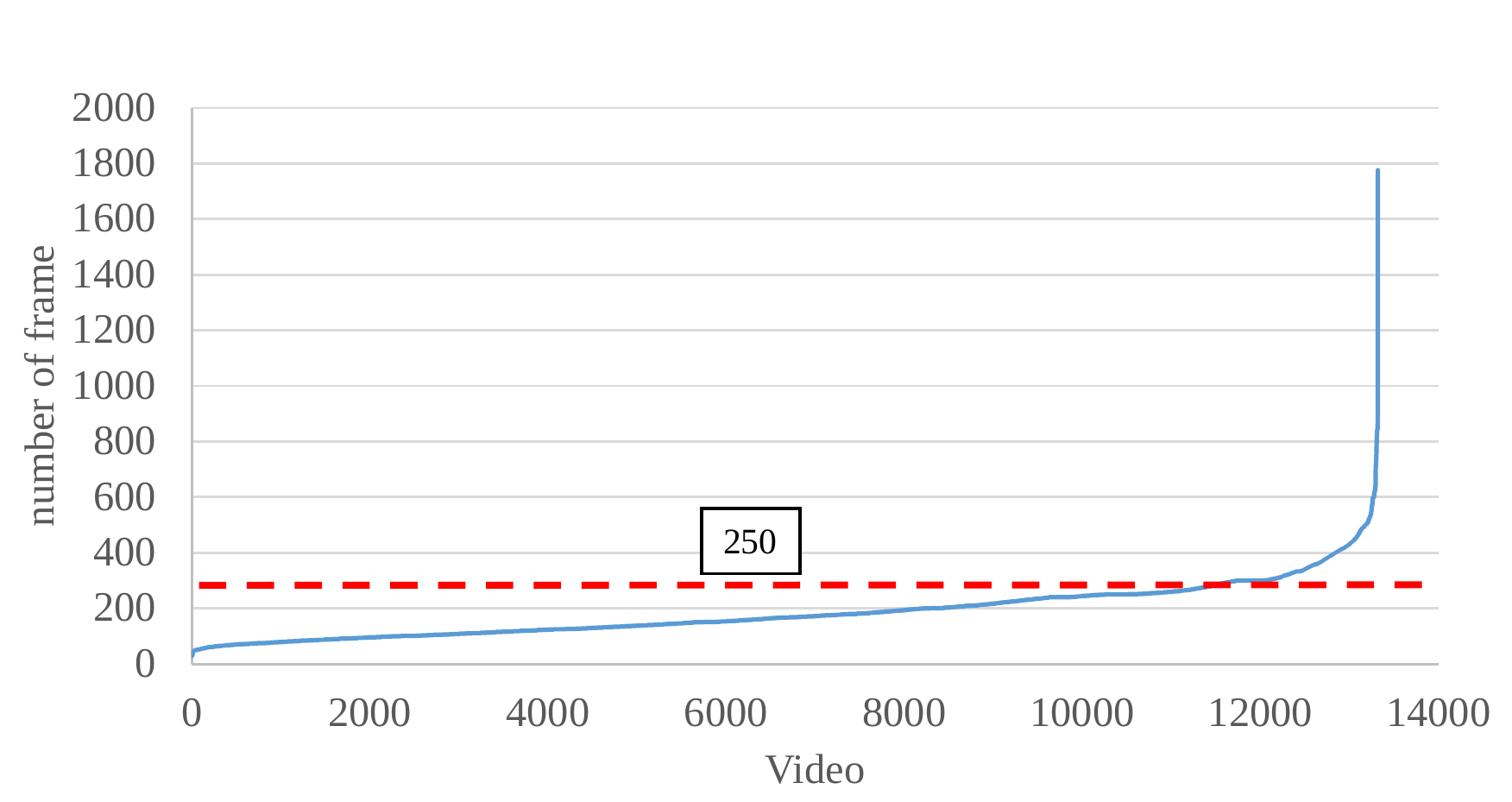}   
    \caption{Statistics of video length of the UCF101 dataset.}
    \label{fig:ucf101-statics}
\end{figure}

\begin{table*}[!htbp]
\small
\centering
\caption{Complete performance comparison of Temporal Segment LSTM on UCF101 split 1. () denote as stacked LSTM.}
\label{table: temporal-segment-lstm}
\begin{tabular}{ccccccc}
TS &BN      & FC   & Temporal Pooling      & BN      & FC         & Accuracy    \\ \hline
\multicolumn{7}{c}{\textbf{ConvNet + LSTM}} \\ \hline
1  &        &      & LSTM-512              &         & FC-101     & 87.0       \\
1  &        &      & LSTM-1024             &         & FC-101     & 86.2       \\
1  &        &      & LSTM-2048             &         & FC-101     & 85.8        \\ \hline
\multicolumn{7}{c}{\textbf{Batch Normalization + LSTM}} \\ \hline
1  &BN      &      & LSTM-512              & BN      & FC-101     & 92.8       \\
1  &BN      &      & LSTM-1024             & BN      & FC-101     & 91.2       \\
1  &BN      &      & LSTM-2048             & BN      & FC-101     & 91.8       \\ 
1  &BN      &      & LSTM-(512, 512)       & BN      & FC-101     & 92.6       \\ 
1  &BN      &      & LSTM-(1024, 512)      & BN      & FC-101     & 91.9       \\ 
1  & BN     &      & LSTM-(512,512,512)    & BN      & FC-101     & 92.1       \\ \hline

\multicolumn{7}{c}{\textbf{Temporal Segment + Max Pooling + Batch Normalization}} \\ \hline
1  &        &      & Max                   &         & FC-101     & 89.0       \\
1  &        &      & Max                   & BN      & FC-101     & 92.8      \\
1  &BN      &      & Max                   & BN      & FC-101     & 92.8       \\ 
3  &BN      &      & Max                   & BN      & FC-101     & 93.4      \\ 
5  &BN      &      & Max                   & BN      & FC-101     & 93.2       \\ \hline
\multicolumn{7}{c}{\textbf{Feature integration + Dimension Reduction}} \\ \hline
3  &BN      & 512  & Max                   & BN      & FC-101     & 93.9      \\ 
5  &BN      & 512  & Max                   & BN      & FC-101     & 93.7       \\ \hline
\multicolumn{7}{c}{\textbf{Temporal Segment + Max + LSTM}} \\ \hline
3  &BN      &      & Max + 512             & BN      & FC-101     & 94.3      \\ 
3  &BN      &      & Max + 1024            & BN      & FC-101     & 94.0       \\ 
3  &BN      &      & Max + 2048            & BN      & FC-101     & 94.1      \\ 
5  &BN      &      & Max + 512             & BN      & FC-101     & 94.2       \\ 
5  &BN      &      & Max + 1024            & BN      & FC-101     & 94.1      \\ 
5  &BN      &      & Max + 2048            & BN      & FC-101     & 94.2       \\ \hline
\multicolumn{7}{c}{\textbf{Temporal Segment + Max + stacked LSTM}} \\ \hline
3  &BN      &      & Max + (512, 512)      & BN      & FC-101     & 94.2       \\
3  &BN      &      & Max + (1024, 512)     & BN      & FC-101     & 94.1       \\
3  &BN      &      & Max + (512, 512, 512)      & BN      & FC-101     & 93.9       \\
5  &BN      &      & Max + (512, 512)      & BN      & FC-101     & 94.2       \\ 
5  &BN      &      & Max + (1024, 512)     & BN      & FC-101     & 94.0       \\ 
5  &BN      &      & Max + (512, 512, 512)      & BN      & FC-101     & 93.8       \\ \hline
\end{tabular}
\end{table*}

\subsection{Complete Experimental Results for Temporal-ConvNet}
We claim that by properly leveraging temporal information, we can achieve state-of-the-art results only using feature vector representations. Table~\ref{table:temporal-convnet} shows all of the architectures with different designs of Temporal-ConvNet layers (\textit{TCL}s). There are several interesting findings: First, the Multi-flow architecture does not guarantee better performance. If we only apply single or double \textit{TCL}, the overall dimension number is still large and can cause over-fitting problems. Applying the multi-flow modules will reduce the performance. However, Temporal-Inception uses four \textit{TCL}s to reduce the temporal dimension to avoid the over-fitting problem, so the performance boosts as expected. Secondly, different multi-flow approaches also affect the results. Temporal-Inception fuses the outputs of flows in each layer instead of fusing at last as Temporal-VGG does. In this way, the architecture can effectively exploit and combine temporal information obtained through various temporal receptive fields, and properly increases the accuracy. Finally, the dimension of last full-connected layer also play an important role. To meet the balance between the capability of discrimination and over-fitting, we choose 1024 as our final feature dimension.

\begin{table*}[!htbp]
\small
\centering
\caption{Complete Performance of different architectures in Temporal-ConvNet on UCF101 split 1. 1L: single \textit{TCL}. 2L: double \textit{TCL}. BN: Batch Normalization. FC: fully-connected layer. In the column ``Architecture", ``T" is denoted as one \textit{TCL}. \{\} denote as the stacked architecture. () denotes as the wide (parallel) architecture. }
\label{table:temporal-convnet}
\begin{tabular}{ccccc}
Architecture            & BN    & Dropout & \multicolumn{1}{c}{FC}      & Accuracy    \\ \hline
\multicolumn{5}{c}{\textbf{1L}} \\ \hline
T                   & BN    & Dropout & \multicolumn{1}{c|}{1024}   & 93.6    \\  
(T,T)               & BN    & Dropout & \multicolumn{1}{c|}{1024}   & 92.6    \\  \hline
\multicolumn{5}{c}{\textbf{2L}} \\ \hline
\{T,T\}             & BN    & Dropout & \multicolumn{1}{c|}{1024}     & 93.1    \\  
\{(T,T),(T,T)\}     & BN    & Dropout & \multicolumn{1}{c|}{1024}     & 92.7    \\  \hline
\multicolumn{5}{c}{\textbf{Temporal-VGG}} \\ \hline
\multicolumn{1}{c|}{}               & BN    & Dropout& \multicolumn{1}{c|}{512}   & 93.5    \\  
\multicolumn{1}{c|}{\{T,T,T,T\}}              & BN    & Dropout& \multicolumn{1}{c|}{1024}    & 94.0    \\  
\multicolumn{1}{c|}{}               & BN    & Dropout& \multicolumn{1}{c|}{2048}    & 93.8    \\  
\multicolumn{1}{c|}{}               & BN    & Dropout& \multicolumn{1}{c|}{4096}    & 93.2    \\  \hline
\multicolumn{5}{c}{\textbf{Multi-flow Temporal-VGG}} \\ \hline  
\multicolumn{1}{c|}{}     & BN    & Dropout & \multicolumn{1}{c|}{512}    & 93.6    \\  
\multicolumn{1}{c|}{(\{T,T,T,T\},\{T,T,T,T\})}      & BN    & Dropout & \multicolumn{1}{c|}{1024}   & 93.4    \\  
\multicolumn{1}{c|}{}     & BN    & Dropout & \multicolumn{1}{c|}{2048}   & 93.5    \\  
\multicolumn{1}{c|}{}     & BN    & Dropout & \multicolumn{1}{c|}{4096}   & 93.2    \\  \hline
\multicolumn{5}{c}{\textbf{Temporal-Inception}} \\ \hline
\multicolumn{1}{c|}{} &     &         & \multicolumn{1}{c|}{1024}   & 93.3              \\
\multicolumn{1}{c|}{} &    & Dropout  & \multicolumn{1}{c|}{1024}   & 93.4    \\       
\multicolumn{1}{c|}{}   & BN    &     & \multicolumn{1}{c|}{1024}   & 93.4    \\
\multicolumn{1}{c|}{\{(T,T),(T,T),(T,T),(T,T)\}} & BN & Dropout & \multicolumn{1}{c|}{}   & 92.7    \\
\multicolumn{1}{c|}{}   & BN & Dropout  & \multicolumn{1}{c|}{512}    & 93.1    \\
\multicolumn{1}{c|}{}   & BN & Dropout  & \multicolumn{1}{c|}{1024}     & \textbf{94.2} \\
\multicolumn{1}{c|}{}   & BN & Dropout  & \multicolumn{1}{c|}{2048}   & 93.7    \\
\multicolumn{1}{c|}{}   & BN & Dropout  & \multicolumn{1}{c|}{4096}   & 93.4            
\\ \hline
\end{tabular}
\end{table*}

Although \textit{TCL}s can reduce the temporal dimension, the filter dimension will increase because of the concatenation of multi-flow modules. There are two different approaches to reduce the filter dimension: (i) reduce the dimension for each multi-flow module (ii) reduce the dimension after all the multi-flow modules. Average pooling, max pooling and Conv fusion (convolve with a set of filters to reduce the filter dimension.) are used as the dimension-reduction methods. Table \ref{table:tem-conv-reduceDim} shows the experiment results. The accuracy drops dramatically if applying the above methods for each module because we partially lose information for each dimension-reduction process. We also found that using multiple convolution layers to gradually reduce the dimension is better than directly applying average or max pooling. 

\begin{table*}[!htbp]
\small
\centering
\caption{Dimension reduction methods for Temporal-ConvNet. The performance is shown on UCF101 split1. Conv1, n: convolution with the kernel size 1$\times$1 and the output filter dimension is n.}
\label{table:tem-conv-reduceDim}
\begin{tabular}{ccccccc}
\multicolumn{7}{c}{\textbf{Reduce the dimension for each multi-flow module}} \\ \hline
\multicolumn{6}{c|}{Average pooling}        & 92.8    \\  
\multicolumn{6}{c|}{Max pooling}                & 92.9    \\  
\multicolumn{6}{c|}{Conv fusion (Conv1, 1)}     & 92.8    \\  \hline

\multicolumn{7}{c}{\textbf{Reduce the dimension after all the multi-flow modules}} \\ \hline
  & Average pooling & & & & \multicolumn{1}{c|}{}   &   93.0  \\  
  & Max pooling   & & & & \multicolumn{1}{c|}{}   &   93.1  \\  
  & Conv fusion   & Conv1, 1 &  & & \multicolumn{1}{c|}{}   & 92.8    \\  
  & Conv fusion   & Conv1, 2 & Conv1, 1 & & \multicolumn{1}{c|}{}   & 92.9    \\  
  & Conv fusion   & Conv1, 4 & Conv1, 1 & & \multicolumn{1}{c|}{}   & 92.5    \\  
  & Conv fusion   & Conv1, 8 & Conv1, 1 & & \multicolumn{1}{c|}{}   & 93.3    \\  
  & Conv fusion   & Conv1, 4 & Conv1, 2 & Conv1, 1 & \multicolumn{1}{c|}{}    & \textbf{94.2}   \\  
  & Conv fusion   & Conv1, 8 & Conv1, 4 & Conv1, 1 & \multicolumn{1}{c|}{}    & 93.8    \\  
  & Conv fusion   & Conv1, 8 & Conv1, 2 & Conv1, 1 & \multicolumn{1}{c|}{}    & 92.7    \\  
  & Conv fusion   & Conv1, 8 & Conv1, 4 & Conv1, 2 &  \multicolumn{1}{c|}{Conv1, 1}   & 93.0    \\  \hline
\end{tabular}
\end{table*}

One of the most important components in the Temporal-ConvNet is the \textit{TCL}, and the convolutional kernel size is the most critical part since it directly affects how the network learns the temporal correlation. Larger kernel sizes are used for actions with longer temporal duration. In our experiments, combining the Temporal-Inception architecture with the convolution kernel size 5 and 7 provides the best capability to represent different kinds of actions. Table \ref{table:tem-conv-conv} also shows the results for other kernel sizes. We also found that the factorization concept in Inception-v3\cite{szegedy2015inceptionv2} does not fit our architecture. Finally, in addition to one-stride convolution followed by max pooling, two-stride convolution could be a better alternative to reduce the dimension since we may lose part of the information by max pooling. However, it is not the case for the Temporal-Inception. Although 2-stride convolution improves the performance on split 1, the overall accuracy is still not better.

\begin{table*}[!htbp]
\centering
\caption{Convolution methods for Temporal-ConvNet. The performance is shown on UCF101 split1. Conv1, n: convolution with the kernel size 1$\times$1 and the output filter dimension is n.}
\label{table:tem-conv-conv}
\begin{tabular}{cccc}
Architecture  & First flow  & \multicolumn{1}{c}{Second flow}       & Accuracy    \\ \hline
1-stride Conv + Max pooling & Conv3, 1 & \multicolumn{1}{c|}{Conv5, 1}    & 92.0    \\  
1-stride Conv + Max pooling & Conv3, 1 & \multicolumn{1}{c|}{Conv7, 1}    & 93.1    \\  
1-stride Conv + Max pooling & Conv3, 1 & \multicolumn{1}{c|}{Conv9, 1}    & 93.8    \\  
1-stride Conv + Max pooling & Conv5, 1 & \multicolumn{1}{c|}{Conv9, 1}    & 92.8    \\  
1-stride Conv + Max pooling & Conv7, 1 & \multicolumn{1}{c|}{Conv9, 1}    & 93.9    \\  
1-stride Conv + Max pooling & Conv5, 1 & \multicolumn{1}{c|}{Conv7, 1}    & \textbf{94.2} (3 splits: 93.9)    \\  
\hline
\multicolumn{4}{c}{\textbf{stacked Conv3 to replace Conv5 \& Conv7}} \\ 
\hline
1-stride Conv + Max pooling & Conv3, 1 - Conv3, 1 & \multicolumn{1}{c|}{Conv3, 1 - Conv3, 1 - Conv3, 1}   & 93.3    \\  
\hline
\multicolumn{4}{c}{\textbf{Use 2-stride Conv to reduce the temporal dimension
}} \\ 
\hline
2-stride Conv & Conv5, 1 & \multicolumn{1}{c|}{Conv7, 1}    & 94.4 (3 splits: 93.6)   \\  
\hline
\end{tabular}
\end{table*}

\subsection{Statistics of the UCF101 dataset}
Fig. \ref{fig:ucf101-statics} depicts the length of each video from the UCF101 dataset. We performed an experiment where we limited the baseline and the two proposed methods to only access the first 10 seconds of each video, e.g. maximum 250 frames per video. Out of 13320 videos in UCF101, there are 2805 videos that are longer than 250 frames. About 21\% of the videos are cut short to demonstrate how the length of videos affects final prediction accuracy. Our experiment shows that the proposed methods can effectively leverage the temporal information since performance can continue to improve as they process additional temporal data.

\subsection{Video Analysis of the UCF101 dataset}
\begin{figure*}[!htbp]
    \centering
    \begin{subfigure}[b]{0.25\textwidth}
        \includegraphics[width=\linewidth]{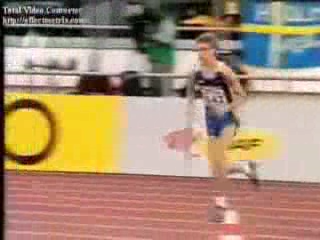}
        \caption{HighJump}
        \label{fig:highjump-run1}
    \end{subfigure}%
  \begin{subfigure}[b]{0.25\textwidth}
        \includegraphics[width=\linewidth]{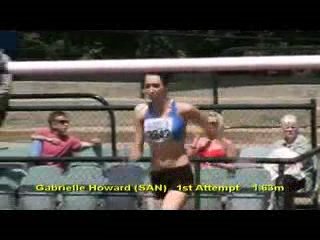}
        \caption{HighJump}
        \label{fig:highjump-run2}
    \end{subfigure}%
    \begin{subfigure}[b]{0.25\textwidth}
        \includegraphics[width=\linewidth]{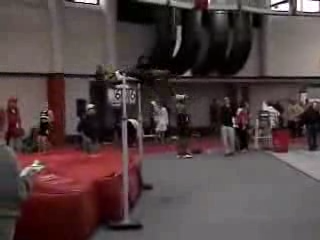}
        \caption{HighJump}
        \label{fig:highjump-jump1}
    \end{subfigure}%
    \begin{subfigure}[b]{0.25\textwidth}
        \includegraphics[width=\linewidth]{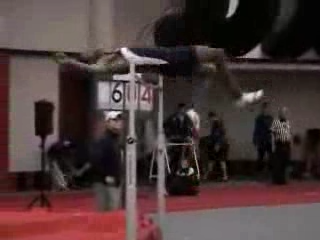}
        \caption{HighJump}
        \label{fig:highjump-jump2}
    \end{subfigure}
    
    \begin{subfigure}[b]{0.25\textwidth}
        \includegraphics[width=\linewidth]{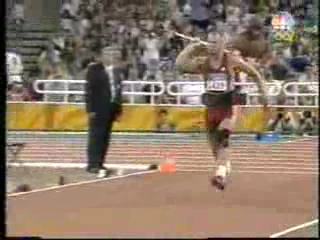}
        \caption{JavelinThrow}
        \label{fig:javelinThrow}
    \end{subfigure}%
    \begin{subfigure}[b]{0.25\textwidth}
        \includegraphics[width=\linewidth]{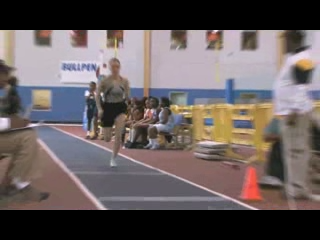}
        \caption{LongJump}
        \label{fig:longjump}
  \end{subfigure}%
    \begin{subfigure}[b]{0.25\textwidth}
        \includegraphics[width=\linewidth]{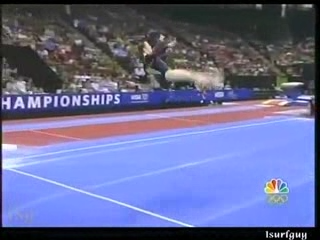}
        \caption{FloorGymnastics}
        \label{fig:floorgymnastics}
    \end{subfigure}%
    \begin{subfigure}[b]{0.25\textwidth}
        \includegraphics[width=\linewidth]{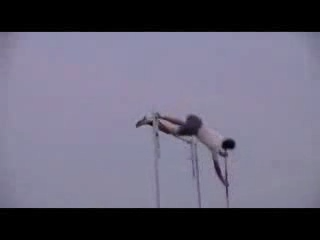}
        \caption{PoleVault}
        \label{fig:polevault}
    \end{subfigure}
    \caption{\textbf{The category (HighJump) that is misclassified by the baseline approach but correctly classified by TS-LSTM and Temporal-Inception.} 1st row: four example videos with the ground truth labels. 2nd row: the incorrectly predicted labels by the baseline approach and the examples of those labels. }
    \label{fig:highjump}
\end{figure*}

We use some examples to show how our approaches work. Our first example is \textit{HighJump}. All the videos in this category can be divided into two parts: running and jumping. The baseline is a frame-based method, which does not make the prediction using the cross-frame information. Therefore, by applying the baseline approach, some videos are misclassified as other categories including running and jumping. For example, Figure~\ref{fig:highjump}(a)(b) are misclassified as \textit{JavelinThrow} and \textit{LongJump}, and Figure~\ref{fig:highjump}(c)(d) are misclassified as \textit{FloorGymnastics} and \textit{PoleVault}. However, those examples are correctly classified by TS-LSTM and Temporal-Inception. This shows both our approaches can effectively extract the temporal information.

\begin{figure*}[!htbp]
    \centering
    \begin{subfigure}[b]{0.25\textwidth}
        \includegraphics[width=\linewidth]{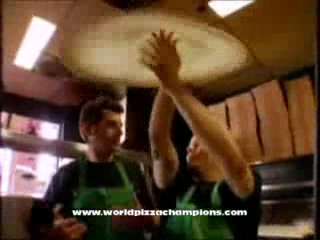}
        \caption{PizzaTossing (type 1)}
        \label{fig:pizzatossing-1}
    \end{subfigure}%
    \begin{subfigure}[b]{0.25\textwidth}
        \includegraphics[width=\linewidth]{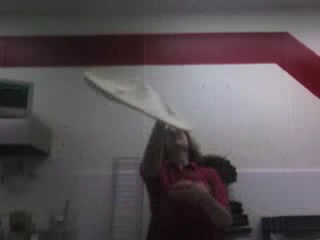}
        \caption{PizzaTossing (type 2)}
        \label{fig:pizzatossing-2}
    \end{subfigure}%
    \begin{subfigure}[b]{0.25\textwidth}
        \includegraphics[width=\linewidth]{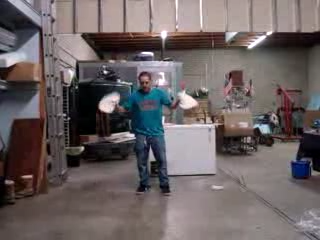}
        \caption{PizzaTossing (type 3)}
        \label{fig:pizzatossing-3}
    \end{subfigure}
    
    \begin{subfigure}[b]{0.25\textwidth}
        \includegraphics[width=\linewidth]{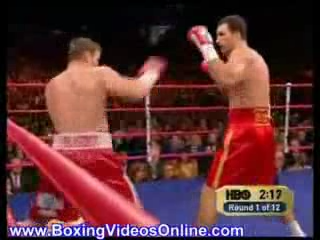}
        \caption{Punch}
        \label{fig:punch}
    \end{subfigure}%
        \begin{subfigure}[b]{0.25\textwidth}
        \includegraphics[width=\linewidth]{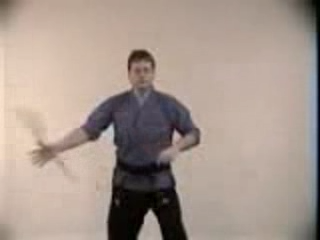}
        \caption{Nunchucks}
        \label{fig:nunchucks}
    \end{subfigure}%
    \begin{subfigure}[b]{0.25\textwidth}
        \includegraphics[width=\linewidth]{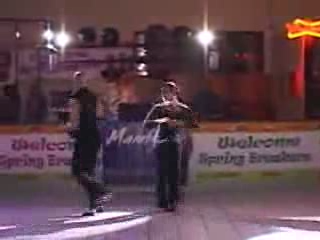}
        \caption{SalsaSpin}
        \label{fig:salsaspin}
    \end{subfigure}
    \caption{\textbf{Another category (PizzaTossing) that is misclassified by the baseline approach but correctly classified by TS-LSTM and Temporal-Inception.} 1st row: 3 example videos with the ground truth labels. 2nd row: the incorrectly predicted labels by the baseline approach and the examples of those labels. }
    \label{fig:pizzatossing}
\end{figure*}

Another example is \textit{PizzaTossing}. In Figure~\ref{fig:pizzatossing}, we show three different variations of this category. Figure~\ref{fig:pizzatossing}(a) is misclassified as \textit{Punch} since both videos include two people and their arm motion are also similar. Figure~\ref{fig:pizzatossing}(b) is misclassified as \textit{Nunchucks} because the people in both videos are making an object spinning. Finally, Figure~\ref{fig:pizzatossing}(c) is misclassified as \textit{SalsaSpin} since in both videos, there are two objects spinning quickly. Those three examples are also correctly classified by both of our approaches because we take the spatial and temporal correlation into account to distinguish different categories.

\subsection{t-SNE Visualization}

\begin{figure*}[!htbp]
    \centering
    \begin{subfigure}[b]{0.33\textwidth}
        \includegraphics[width=\linewidth]{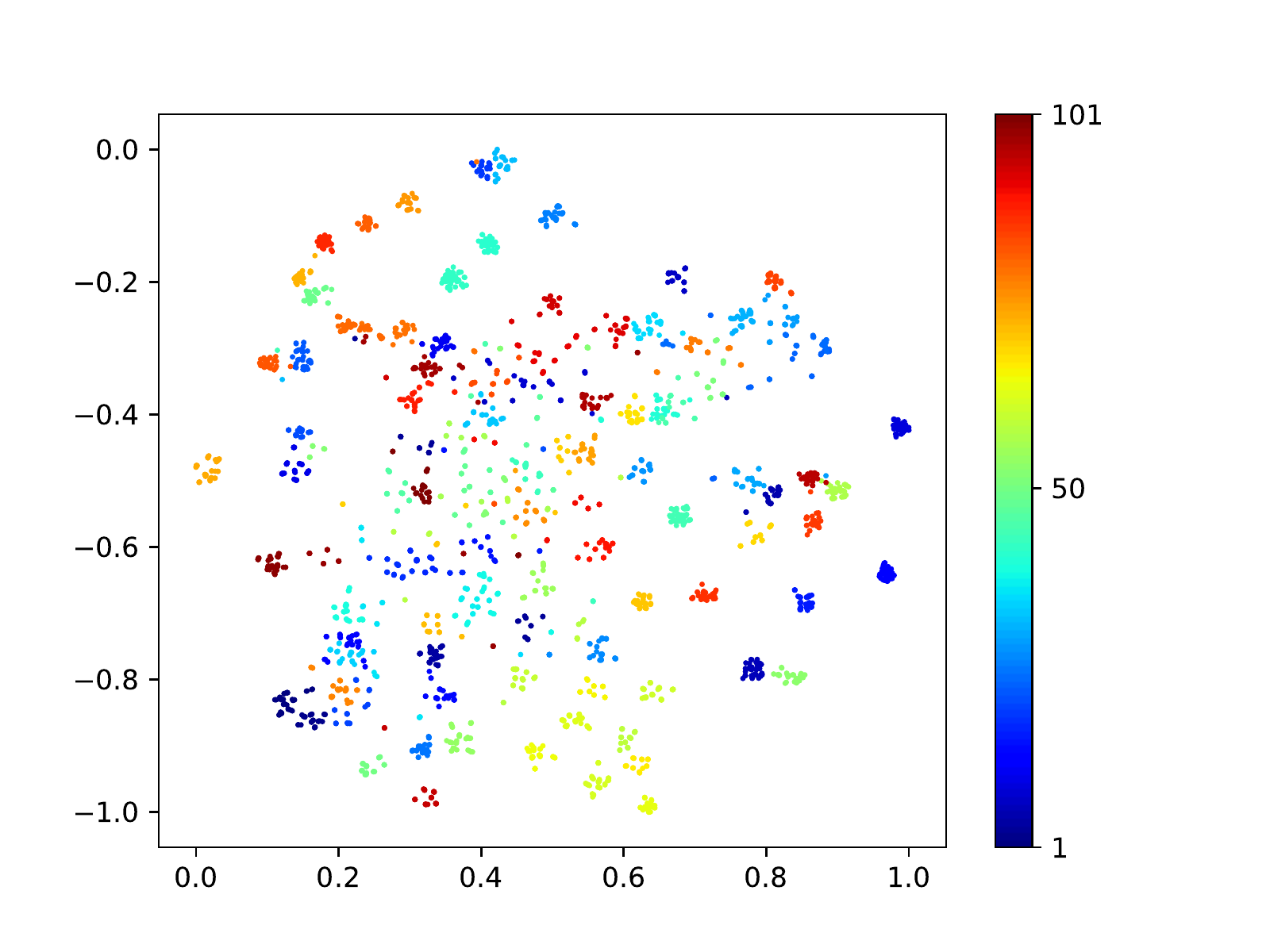}
        \caption{Two-stream ConvNet}
        \label{fig:tsne-twostream}
    \end{subfigure}%
    \begin{subfigure}[b]{0.33\textwidth}
        \includegraphics[width=\linewidth]{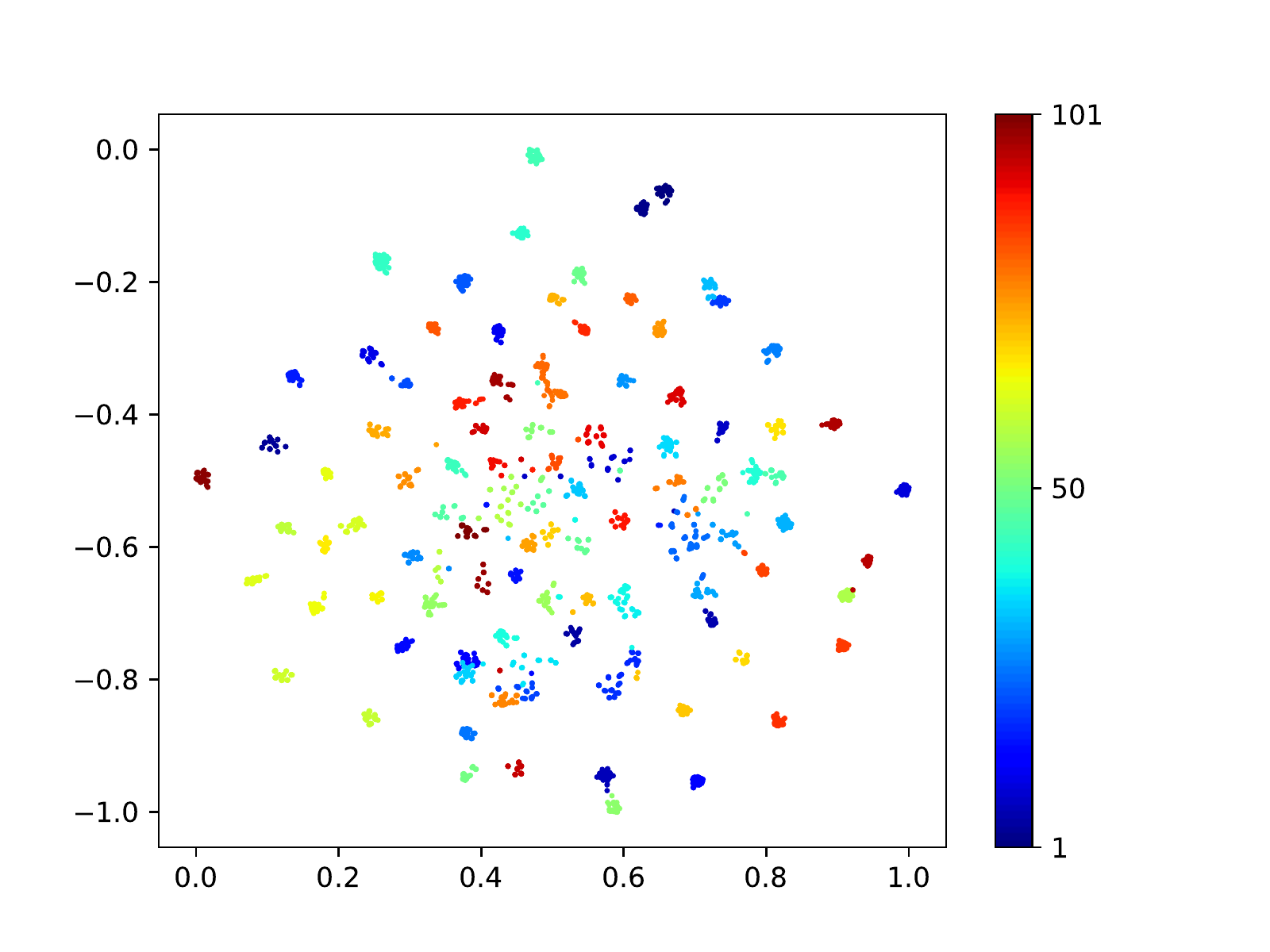}
        \caption{Temporal Segment LSTM}
        \label{fig:tsne-tslstm}
    \end{subfigure}%
    \begin{subfigure}[b]{0.33\textwidth}
        \includegraphics[width=\linewidth]{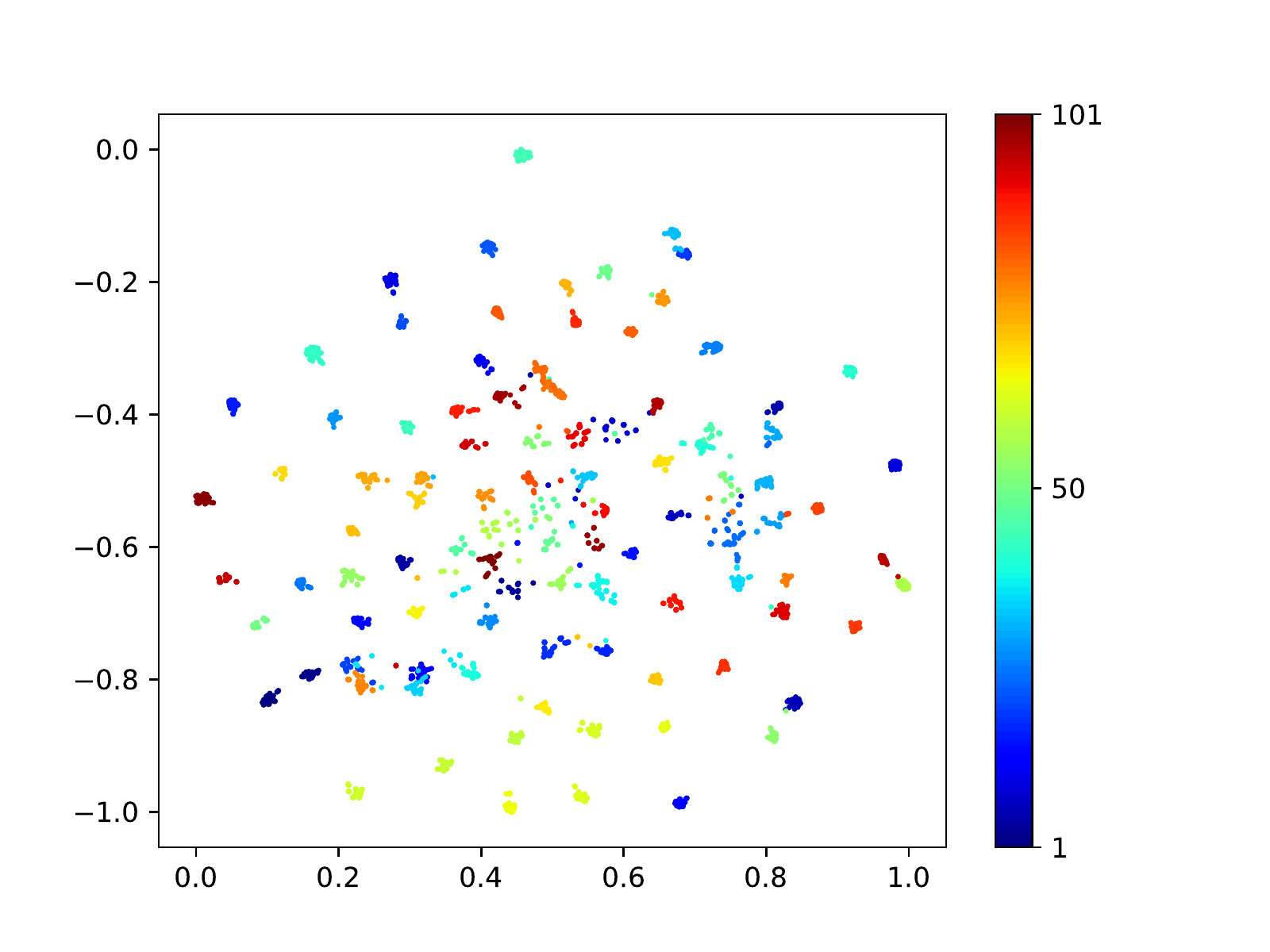}
        \caption{Temporal-Inception}
        \label{fig:tsne-tcnn}
    \end{subfigure}
    \caption{t-SNE visualization of the last feature vector representation from baseline two-stream ConvNet, TS-LSTM, and Temporal-Inception on UCF101 split 1. The feature representations from the two proposed methods show the data points are easier to separate and classify.}
    \label{fig:t-SNE}
\end{figure*}

We visualize the feature vectors from baseline two-stream ConvNet, TS-LSTM, and Temporal-Inception methods. From the visualizations in Fig. \ref{fig:t-SNE}, we can see that both of the proposed methods are able to group the test samples into more distinct clusters. Thus, after the last fully-connected layer, both the proposed methods can achieve better classification results.  

\begin{figure*}[!htbp]
    \centering
    \begin{subfigure}[b]{0.3\textwidth}
        \includegraphics[width=\linewidth]{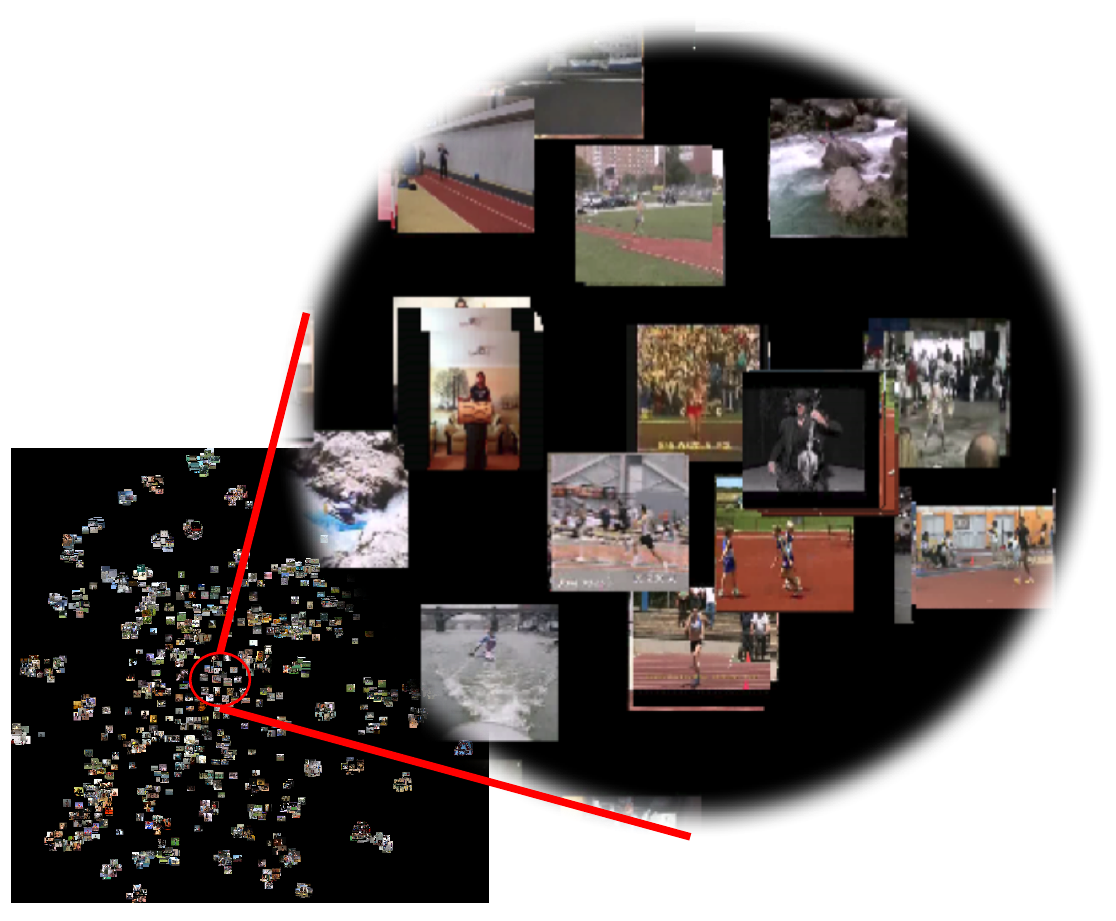}
        \caption{Baseline two-stream ConvNet}
        \label{fig:tsne-baseline-twostream-zoomin}
    \end{subfigure}%
    \begin{subfigure}[b]{0.3\textwidth}
        \includegraphics[width=\linewidth]{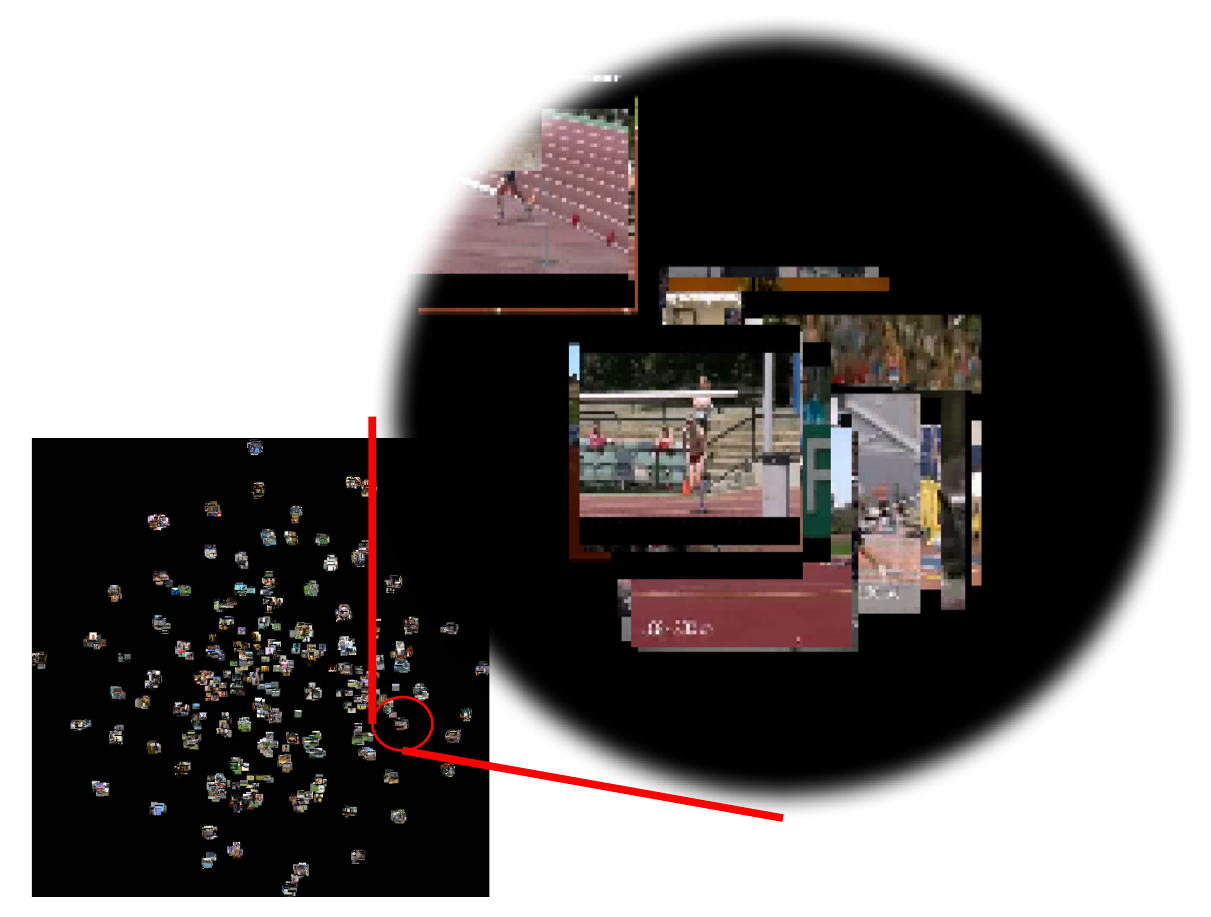}
        \caption{TS-LSTM}
        \label{fig:tsne-tslstm-zoomin}
    \end{subfigure}%
    \begin{subfigure}[b]{0.3\textwidth}
        \includegraphics[width=\linewidth]{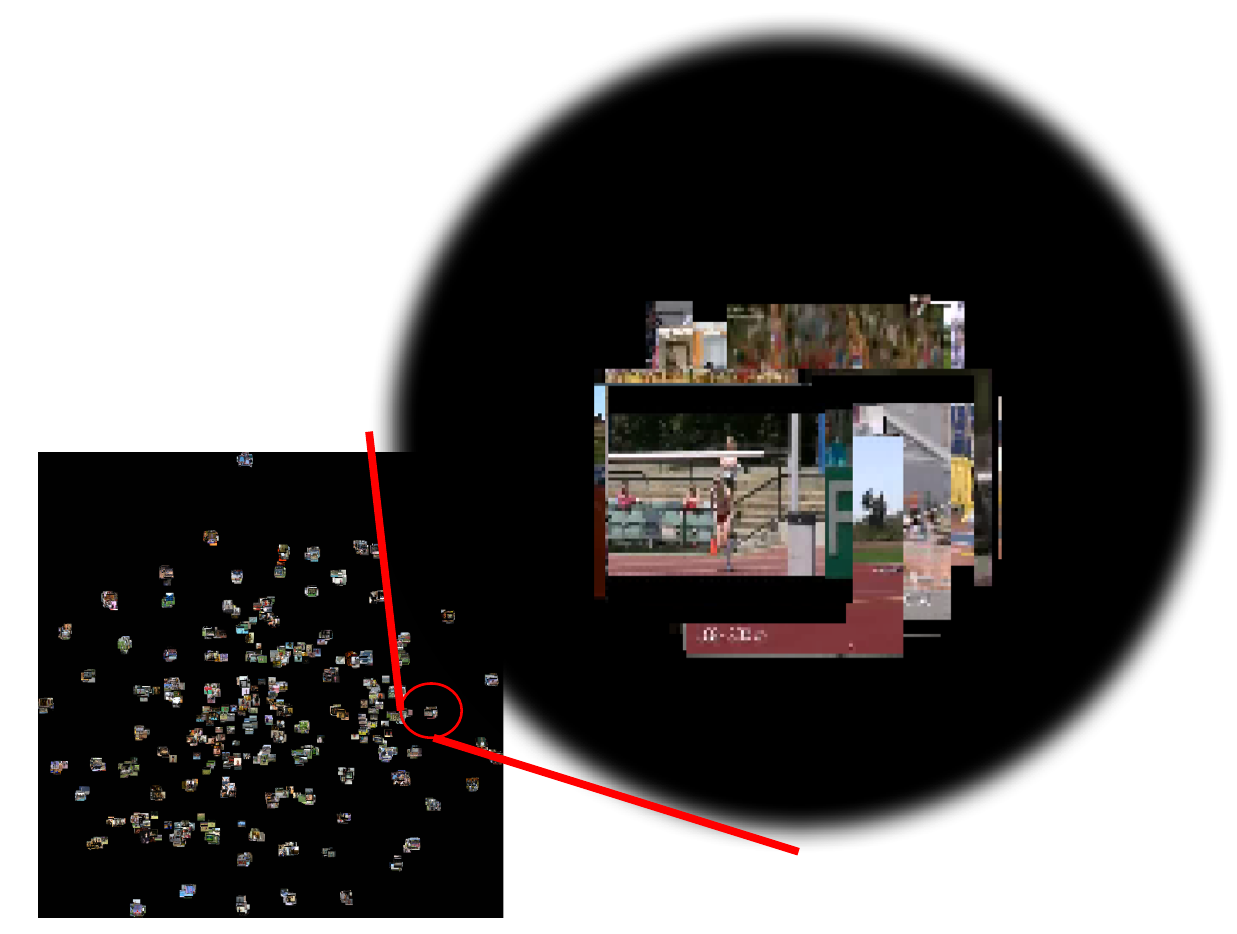}
        \caption{Temporal-Inception}
        \label{fig:tsne-tcnn-zoomin}
    \end{subfigure}

    \begin{subfigure}[b]{0.3\textwidth}
        \includegraphics[width=\linewidth]{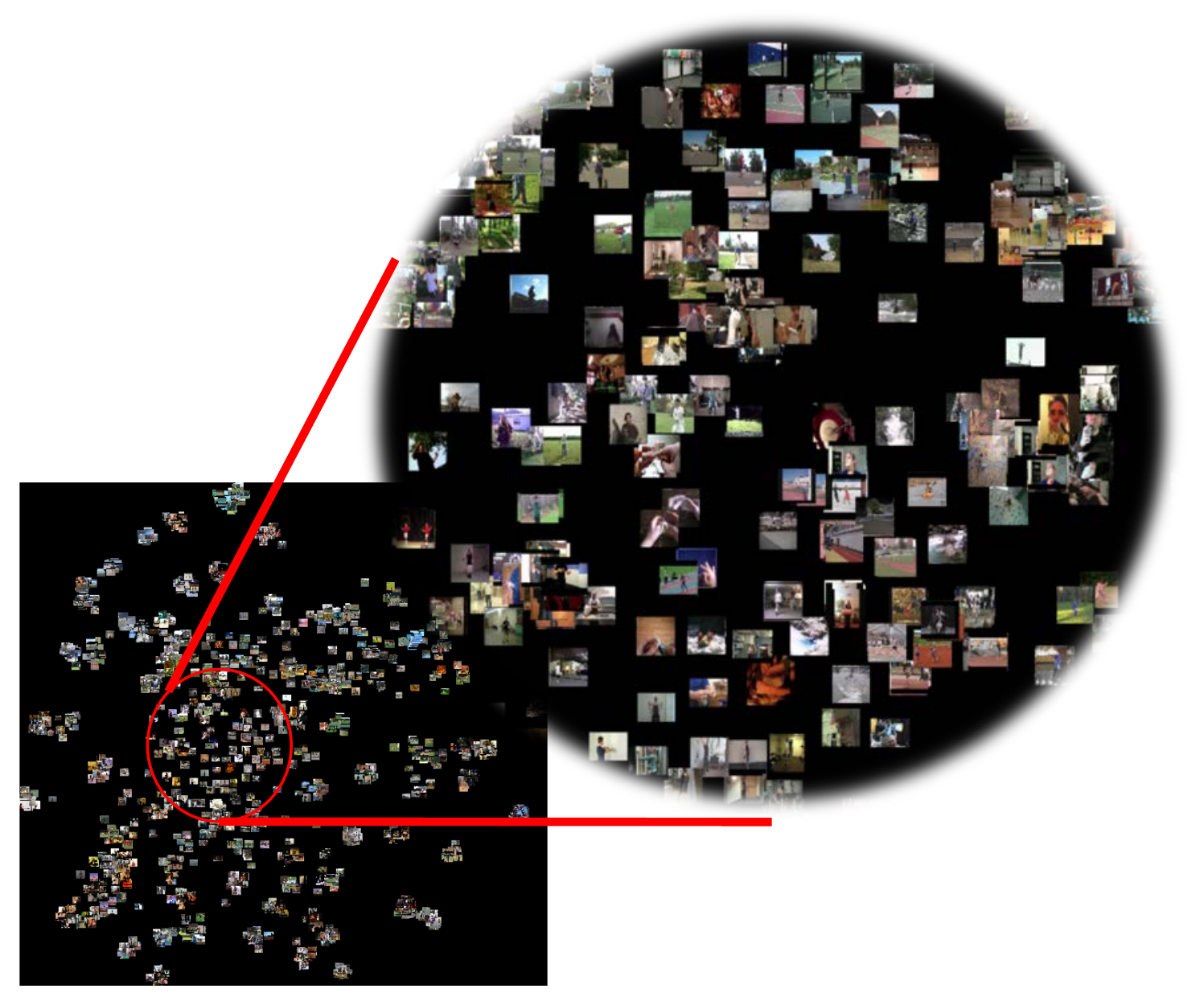}
        \caption{Baseline two-stream ConvNet}
        \label{fig:tsne-baseline-twostream-zoomin-pizza}
    \end{subfigure}%
    \begin{subfigure}[b]{0.3\textwidth}
        \includegraphics[width=\linewidth]{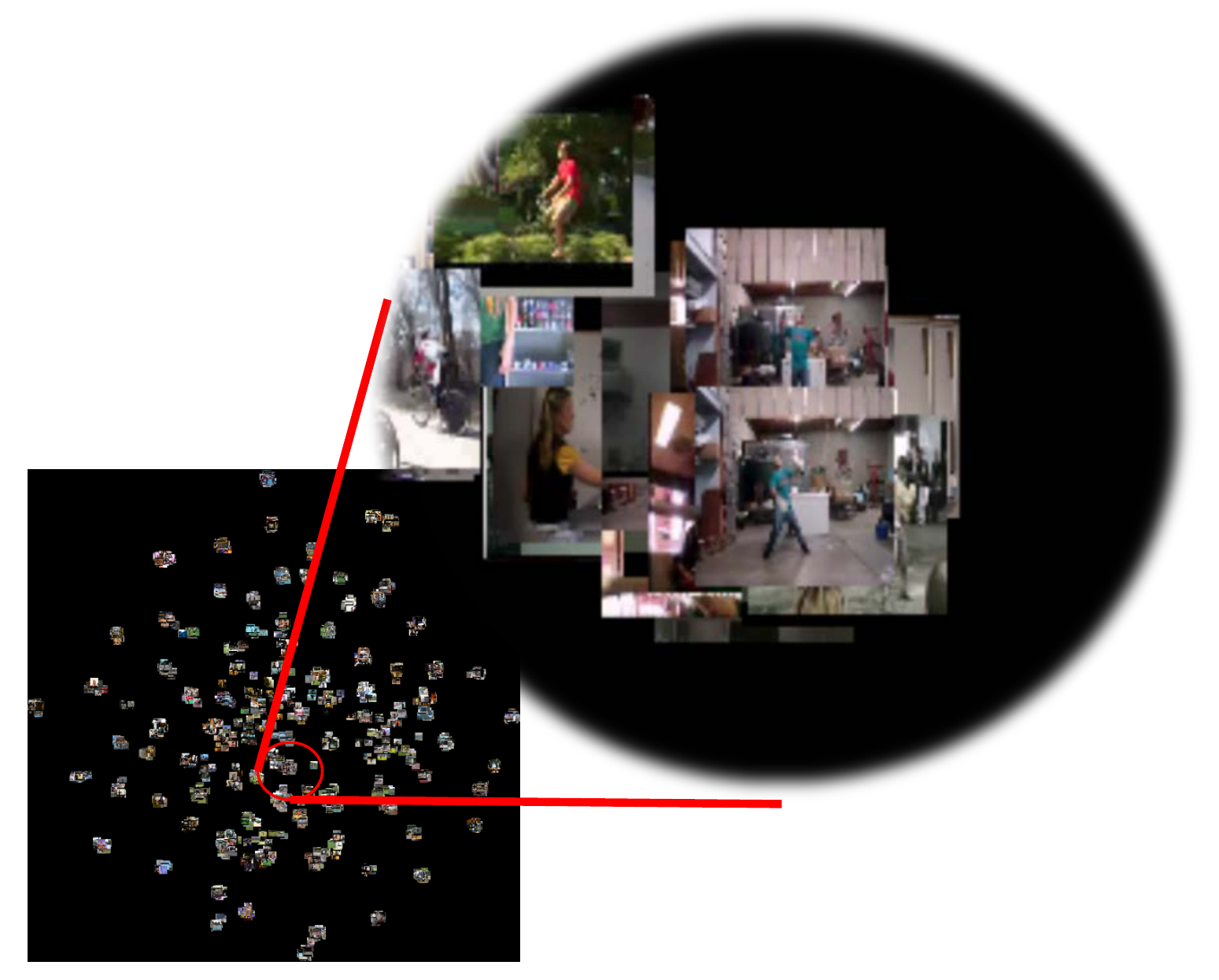}
        \caption{TS-LSTM}
        \label{fig:tsne-tslstm-zoomin-pizza}
    \end{subfigure}%
    \begin{subfigure}[b]{0.3\textwidth}
        \includegraphics[width=\linewidth]{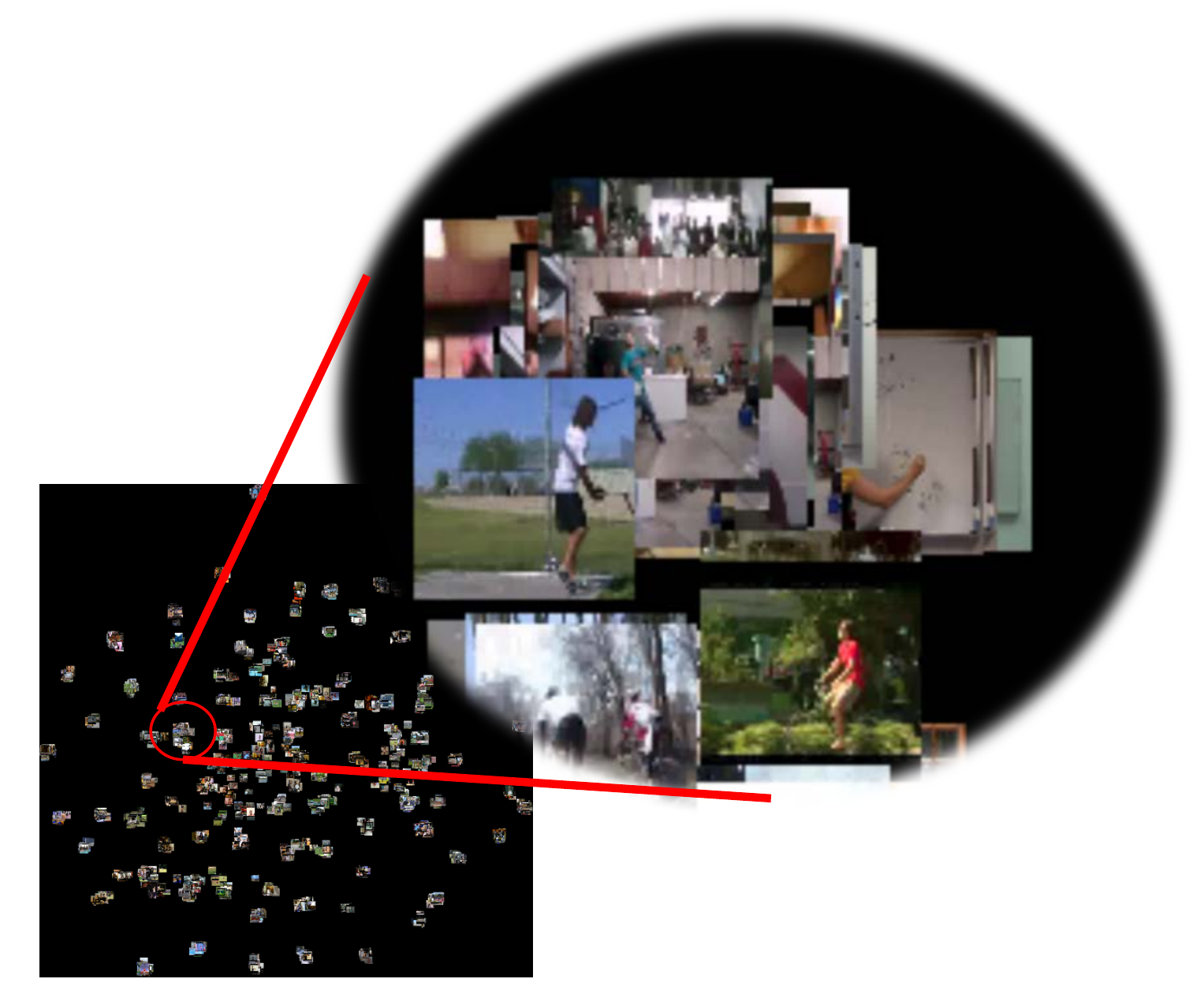}
        \caption{Temporal-Inception}
        \label{fig:tsne-tcnn-zoomin-pizza}
    \end{subfigure}
    \caption{By zooming in the t-SNE visualization of baseline two-stream ConvNet, TS-LSTM, and Temporal-Inception on UCF101 split 1, we can see specific example of how videos in different video classes were scattered before and grouped together by both proposed TS-LSTM and Temporal-Inception methods. Top tow: \textit{HighJump} video class in UCF101. Bottom tow: \textit{PizzaTossing} video class in UCF101. Note that the recognition accuracy of \textit{HighJump} using individual methods are: 62.2\% (baseline), 97.3\% (TS-LSTM), and 94.6\% (Temporal-Inception), and the accuracy of \textit{PizzaTossing} are: 66.7\% (baseline), 90.9\% (TS-LSTM), and 97.0\% (Temporal-Inception)}
    \label{fig:t-SNE-zoomin}
\end{figure*}

\begin{figure}[!htbp]
    \centering
    \includegraphics[width=\linewidth]{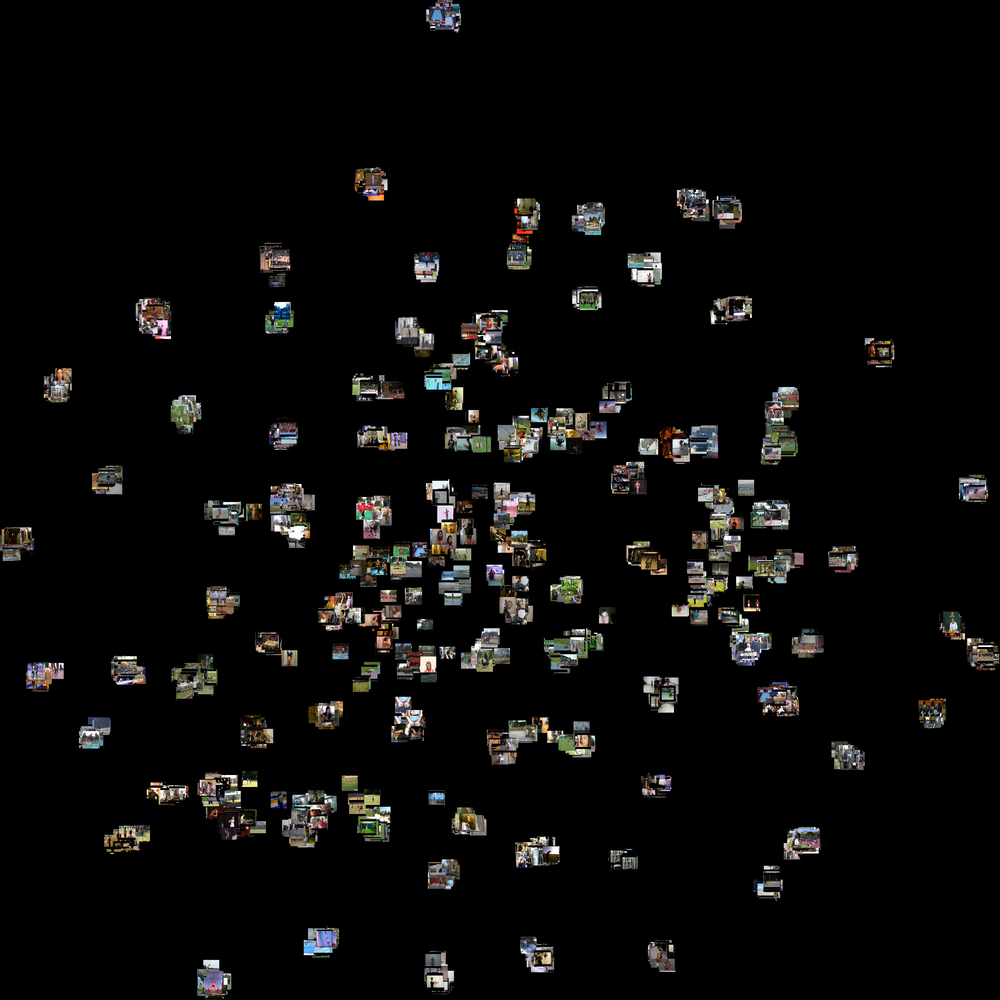}   
    \caption{t-SNE visualization of Temporal-Inception. Each of the data points are replaced by the snap shot of each test video.}
    \label{fig:tsne-temporal-inception}
\end{figure}

By superimpose the snap image from each of the video on the data points optimized from t-SNE, the visualization of the videos in the high-dimensional space are shown in Fig. \ref{fig:tsne-twostream}, \ref{fig:tsne-tslstm}, and \ref{fig:tsne-temporal-inception}. We can further observe the difference from each of the class by zoom-in these figures, as shown in Fig. \ref{fig:t-SNE-zoomin}. It is clear that after using the proposed TS-LSTM and Temporal-Inception, the data points from the same class were pushed toward each other in the high-dimensional space. For example, both \textit{HighJump} and \textit{PizzaTossing} videos were originally scattered around the center of the figures, but were able to grouped together by the two proposed methods. Note that the accuracy of individual methods are: 62.2\% (baseline), 97.3\% (TS-LSTM), and 94.6\% (Temporal-Inception) for \textit{HighJump}; 66.7\% (baseline), 90.9\% (TS-LSTM), and 97.0\% (Temporal-Inception) for \textit{PizzaTossing}.

\begin{figure}[!htbp]
    \centering
    \includegraphics[width=\linewidth]{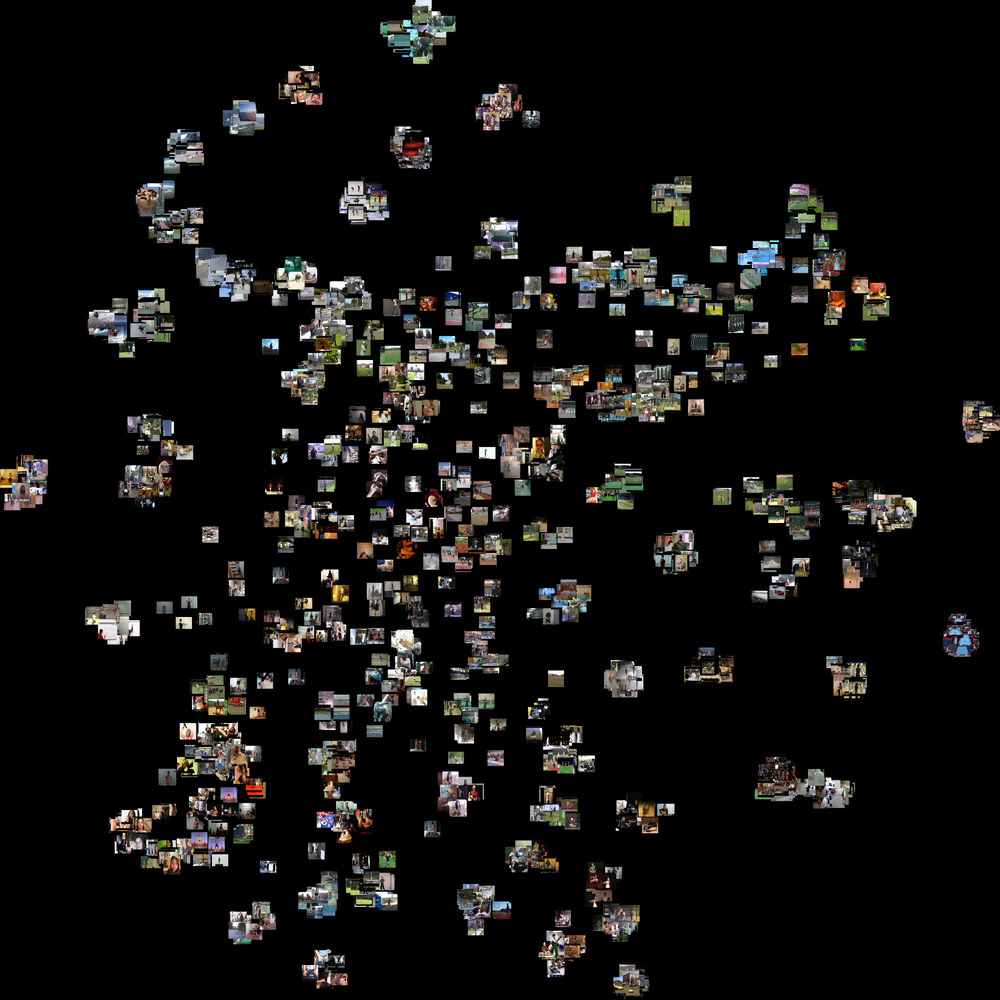}   
    \caption{t-SNE visualization of baseline two-stream ConvNet. Each of the data points are replaced by the snap shot of each test video.}
    \label{fig:tsne-twostream}

    \centering
    \includegraphics[width=\linewidth]{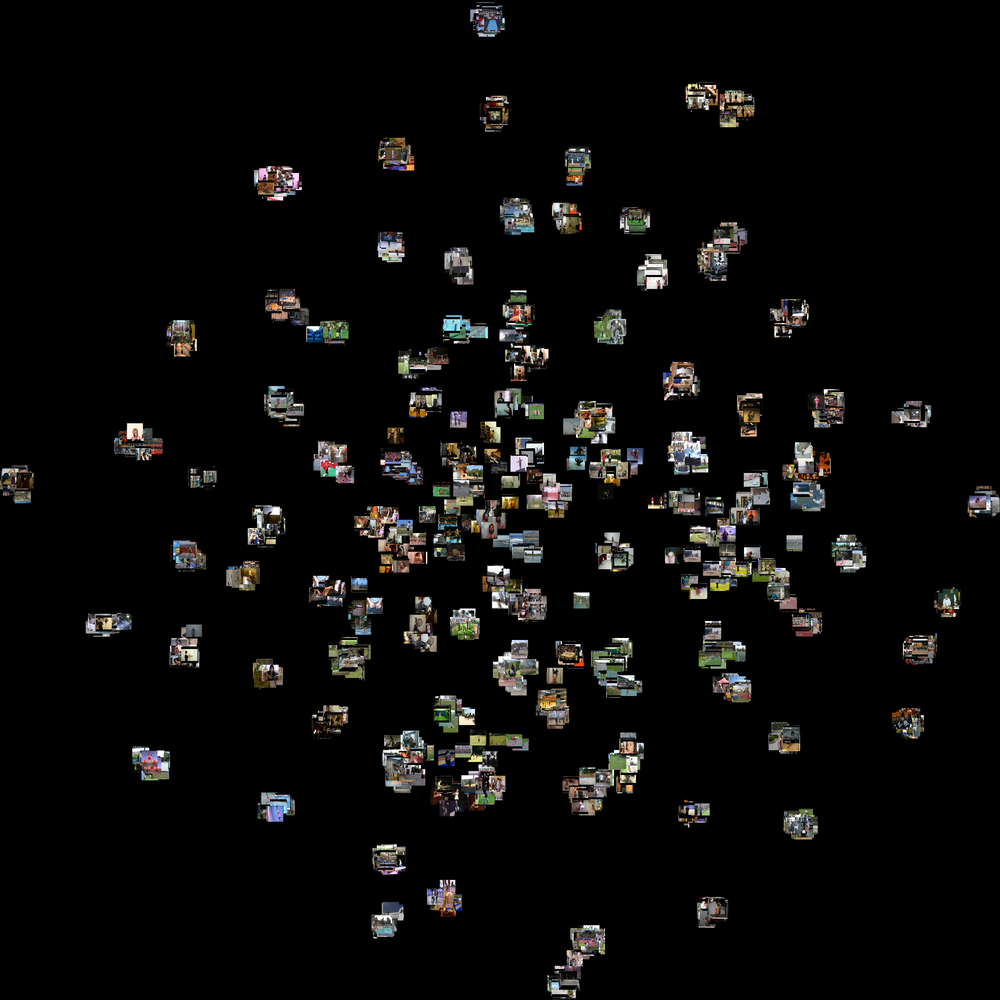}   
    \caption{t-SNE visualization of TS-LSTM. Each of the data points are replaced by the snap shot of each test video.}
    \label{fig:tsne-tslstm}
\end{figure}

{\small
\bibliographystyle{ieee}
\bibliography{egbib}
}

\end{document}